\pdfoutput=1
\documentclass[11pt]{article}

\usepackage[final]{coling}

\usepackage{times}
\usepackage{latexsym}

\usepackage[T1]{fontenc}
\usepackage[utf8]{inputenc}

\usepackage{microtype}

\usepackage{inconsolata}

\usepackage{graphicx}
\usepackage{pifont}
\usepackage{amsmath}
\usepackage{tabularx}
\usepackage{makecell}
\usepackage{array}
\usepackage{colortbl}
\usepackage{booktabs}
\usepackage{multirow}
\usepackage{tabularx}
\usepackage{graphicx}
\usepackage{caption}
\newcolumntype{m}[1]{>{\centering\arraybackslash}p{#1}}
\usepackage{authblk}
\usepackage{enumitem}
\usepackage{subfigure}
\title{KG-FPQ: Evaluating Factuality Hallucination in LLMs with Knowledge\\Graph-based False Premise Questions}

 \author{
 \textbf{Yanxu Zhu\textsuperscript{1}},
 \textbf{Jinlin Xiao\textsuperscript{1}},
 \textbf{Yuhang Wang\textsuperscript{1}},
 \textbf{Jitao Sang\textsuperscript{1,2~\thanks{ \hspace{1mm} Corresponding author}}}
 \\
 \textsuperscript{1}Beijing Key Lab of Traffic Data Analysis and Mining, Beijing Jiaotong University\\
 \textsuperscript{2}Peng Cheng Lab,
\\
 \small{
   \textbf{Correspondence:} \href{mailto:yanxuzhu@bjtu.edu.cn}{yanxuzhu@bjtu.edu.cn}, \href{mailto:jinlinxiao@bjtu.edu.cn}{jinlinxiao@bjtu.edu.cn}, \href{mailto:yhangwang@bjtu.edu.cn}{yhangwang@bjtu.edu.cn}, \href{mailto:jtsang@bjtu.edu.cn}{jtsang@bjtu.edu.cn}
 }
}
\begin{document}
\maketitle
\begin{abstract}
Recent studies have demonstrated that large language models (LLMs) are susceptible to being misled by false premise questions (FPQs), leading to errors in factual knowledge, known as factuality hallucination. Existing benchmarks that assess this vulnerability primarily rely on manual construction, resulting in limited size and lack of expandability. In this work, we introduce an automated, scalable pipeline to  create FPQs based on knowledge graphs (KGs). The first step is to modify true triplets extracted from KGs to create false premises. Subsequently, utilizing the state-of-the-art capabilities of GPTs, we generate semantically rich FPQs. Based on the proposed method, we present a comprehensive benchmark, the \textbf{K}nowledge \textbf{G}raph-based \textbf{F}alse \textbf{P}remise \textbf{Q}uestions (\textbf{KG-FPQ}), which contains approximately 178k FPQs across three knowledge domains, at six levels of confusability, and in two task formats. Using KG-FPQ, we conduct extensive evaluations on several representative LLMs and provide valuable insights. The KG-FPQ dataset and code are available at~\url{https://github.com/yanxuzhu/KG-FPQ}.
\end{abstract}

\section{Introduction}

Large Language Models (LLMs)~\citep{zhao2023survey} excel in natural language understanding and generation but often produce texts that deviate from real-world factual knowledge, a problem known as factuality hallucination~\citep{huang2023survey}. This issue restricts their applicability in scenarios requiring high factual accuracy.\par

\begin{figure}[t] 
  \includegraphics[width=\linewidth]{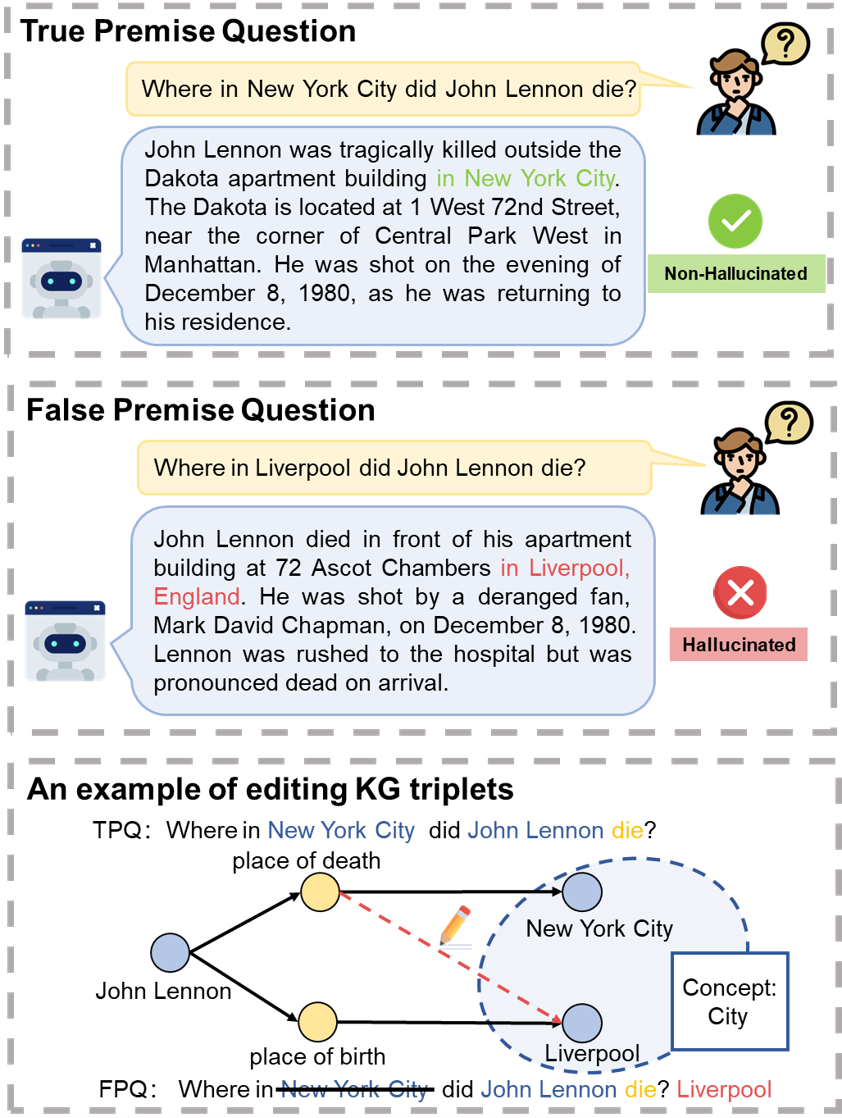}
  \centering
  \caption{Top: LLM correctly answers when faced with a TPQ. Middle: LLM experiences factuality hallucination when faced with an FPQ. Bottom: An example of editing triplets in the KG.} 
  \label{fig:example}
\end{figure}

Recent studies~\citep{vu2023freshllms, yuan2024whispers} have demonstrated that False Premise Questions (FPQs) can induce factuality hallucination in LLMs, as these models often respond directly to FPQs without verifying their validity. An FPQ is a question that contains incorrect facts which are not explicitly stated but might be mistakenly believed by the questioner~\citep{yu-etal-2023-crepe}. For example, as shown at the top of Figure~\ref{fig:example}, when asked with a true premise question (TPQ), the LLM can answer correctly, indicating that the LLM possesses relevant knowledge. However, as depicted in the middle of Figure~\ref{fig:example}, when the TPQ is transformed into an FPQ, the LLM is induced to hallucinate.\par

Before the advent of LLMs such as ChatGPT~\citep{chatgpt2024}, several studies discussed FPQs\citep{yu-etal-2023-crepe, kim-etal-2023-qa, hu-etal-2023-wont}, focusing on the ability of pre-trained language models like RoBERTa~\citep{liu2019robertarobustlyoptimizedbert} and T5~\citep{raffel2023exploringlimitstransferlearning} to detect and correct false premises, rather than addressing the hallucination issue. In the era of LLMs, only a few works have explored the factual hallucination phenomenon induced by FPQs~\citep{vu2023freshllms, yuan2024whispers}. However, ~\citet{vu2023freshllms} rely on a very limited FPQ dataset, and ~\citet{yuan2024whispers} examine a small number of models, resulting in evaluations that are not sufficiently comprehensive and in-depth. Additionally, these studies often depend on manually curated datasets, which limits their scale, expandability, and knowledge coverage.\par

We explore an automated, scalable method to construct FPQs. The first step involves extracting true triplets from knowledge graphs (KGs) and editing them into false triplets. Subsequently, GPTs are utilized to generate FPQs based on these false triplets. Specifically, We extract triplets from a KG in the form of \textit{<subject, relation, object>} and edit the object to create false triplets \textit{<subject, relation, edited object>}. We design editing methods from two perspectives: 1) the edited object at varying distances from the subject in the KG; 2) the edited object having varying associations with the original object in the KG. As the example shown at the bottom of Figure \ref{fig:example}, we edit the true triplet \textit{<John Lennon, place of death, New York City>} to the false triplet \textit{<John Lennon, place of death, Liverpool>}. \textit{Liverpool} is a 1-hop neighbor of \textit{John Lennon} and belongs to the same concept as \textit{New York City} but has a different relation to the subject. There are six editing methods to create false triplets varying in levels of confusability. After editing, we utilize GPT-3.5~\citep{openai2023gpt35} and GPT-4~\citep{openai2024gpt4} to generate FPQs in Yes-No and WH formats respectively corresponding to discriminative and generative evaluation of hallucination~\citep{zhang2023sirens}. By the proposed method, we present a comprehensive benchmark, the \textbf{K}nowledge \textbf{G}raph-based \textbf{F}alse \textbf{P}remise \textbf{Q}uestions, which contains FPQs across three knowledge domains, at six levels of confusability, and in two task formats. The comparison between KG-FPQ and other datasets is detailed in Table~\ref{tab: dataset Comparison}.\par

% Existing FPQ benchmarks \citep{yu-etal-2023-crepe, kim-etal-2023-qa, hu-etal-2023-wont, vu2023freshllms} primarily rely on manual construction, resulting in limited scale and lack of extensibility. \citet{yuan2024whispers} construct their dataset by corrupting triplets in Wikidata \citep{10.1145/2629489} and filling them into human-written templates. However, the dataset covers only two narrow knowledge domains, and the use of fixed templates limits its semantic richness.  Additionally, these studies lack a thorough evaluation of factuality hallucination induced by FPQs. \par

We evaluate the performance of several representative and advanced LLMs on KG-FPQ across both discriminative and generative tasks. Since manual evaluation of the generative task is costly, we introduce an automated evaluator named FPQ-Judge to identify whether responses of LLMs to FPQs are misled by the false premises, achieving a 93\% accuracy rate on a manually annotated test set. Through extensive experiments, we reach three essential conclusions: (1) In terms of confusability, when the edited object has a closer distance with the subject or has a stronger association with the original object, FPQs are more confusing to LLMs. (2) In terms of task formats, LLMs perform worse at generating factual statements than at distinguishing them when faced with FPQs. (3) In terms of knowledge domains, knowledge proficiency of LLMs varies across domains, and there is no positive correlation between knowledge proficiency and the ability to resist the interference of FPQs. Our contributions can be summarized as follows:\par

\begin{itemize}
\item We propose an automated and scalable pipeline combining KGs and GPTs for constructing FPQ datasets, by editing true triplets into false triplets and utilizing GPTs to generate FPQs.

\item Based on the proposed method, we create a comprehensive benchmark, KG-FPQ, containing FPQs across three knowledge domains, at six levels of confusability, and in two task formats.

\item We fine-tune an automated evaluator for generative hallucination evaluation, FPQ-Judge, achieving 93\% accuracy on a manually annotated test set. Furthermore, we conduct an in-depth evaluation of factuality hallucination induced by FPQs on several representative LLMs, yielding valuable insights.
\end{itemize}

\section{Related Work}
\noindent\textbf{Evaluation of Factuality Hallucination}
Many benchmarks evaluate factuality hallucination~\citep{lin2022truthfulqa, li2023halueval,min2023factscore,muhlgay2024generating} due to the risks it poses in practical LLM applications. The evaluation formats are primarily divided into discriminative evaluation~\citep{lin2022truthfulqa,li2023halueval,muhlgay2024generating} and generative evaluation~\citep{lin2022truthfulqa,min2023factscore}, which respectively assess the ability of LLMs to distinguish factual statements and generate factual content~\citep{zhang2023sirens}. Hallucination induced by FPQs belongs to factuality hallucination, and this paper evaluates this vulnerability in both discriminative and generative formats.\\

\begin{table*}[t]
  \centering
  \small
  \begin{tabular}{lccccc}
   \toprule 
   \textbf{Datasets} &  \textbf{Source} & \textbf{Format} & \textbf{Scale} & \textbf{Scalable} & \makecell{\textbf{Varying}\\\textbf{ Confusability}}\\ 
    \midrule
    CREPE~\citep{yu-etal-2023-crepe} & Internet & Gen & $8,400$ & \textcolor{red}{\ding{55}} & \textcolor{red}{\ding{55}}\\
    $\text{(QA)}^2$~\citep{kim-etal-2023-qa} & Internet & Gen & $602$  & \textcolor{red}{\ding{55}}& \textcolor{red}{\ding{55}}\\
    FalseQA~\citep{hu-etal-2023-wont} & Human Written & Gen & $2,365$ & \textcolor{red}{\ding{55}} & \textcolor{green}{\ding{51}}\\
    FRESHQA~\citep{vu2023freshllms} & Human Written & Gen & $600$ & \textcolor{red}{\ding{55}} & \textcolor{red}{\ding{55}}\\
    FAITH~\citep{yuan2024whispers} & KG\&Templates & Gen & $5,832$ & \textcolor{red}{\ding{55}} & \textcolor{red}{\ding{55}}\\
    \midrule
    KG-FPQ(ours) & KG\&LLMs & Dis\&Gen & $14,860 \times 6\times2$ & \textcolor{green}{\ding{51}} & \textcolor{green}{\ding{51}}\\
    \bottomrule
  \end{tabular}
  \caption{Comparison with existing FPQ datasets.} 
  \label{tab: dataset Comparison}
\end{table*} 
\begin{figure*}[t] 
  \includegraphics[width=\linewidth]{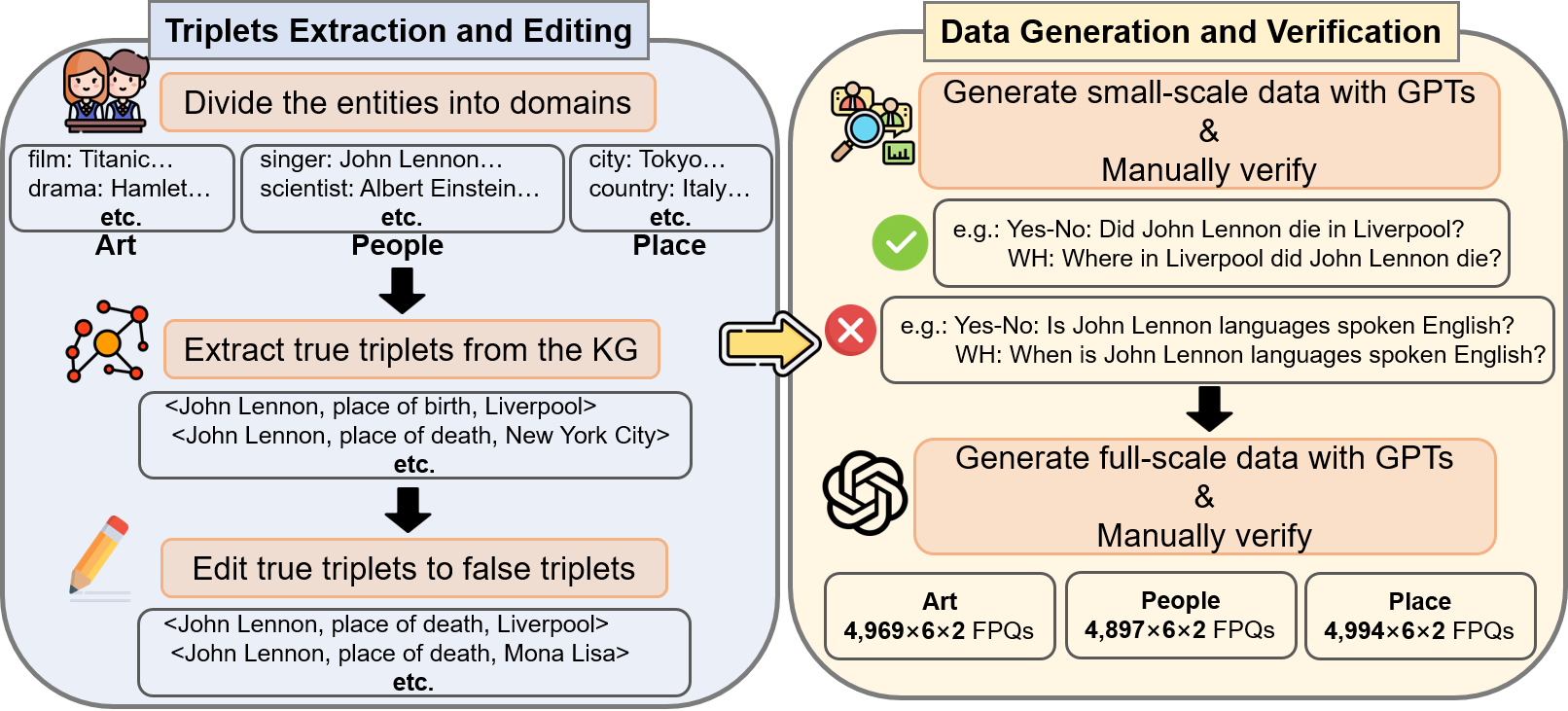}
  \centering
  \caption{Overview of the construction process of KG-FPQ.} 
  \label{fig:pipline}
\end{figure*}

\begin{figure*}[t] 
  \includegraphics[width=\linewidth]{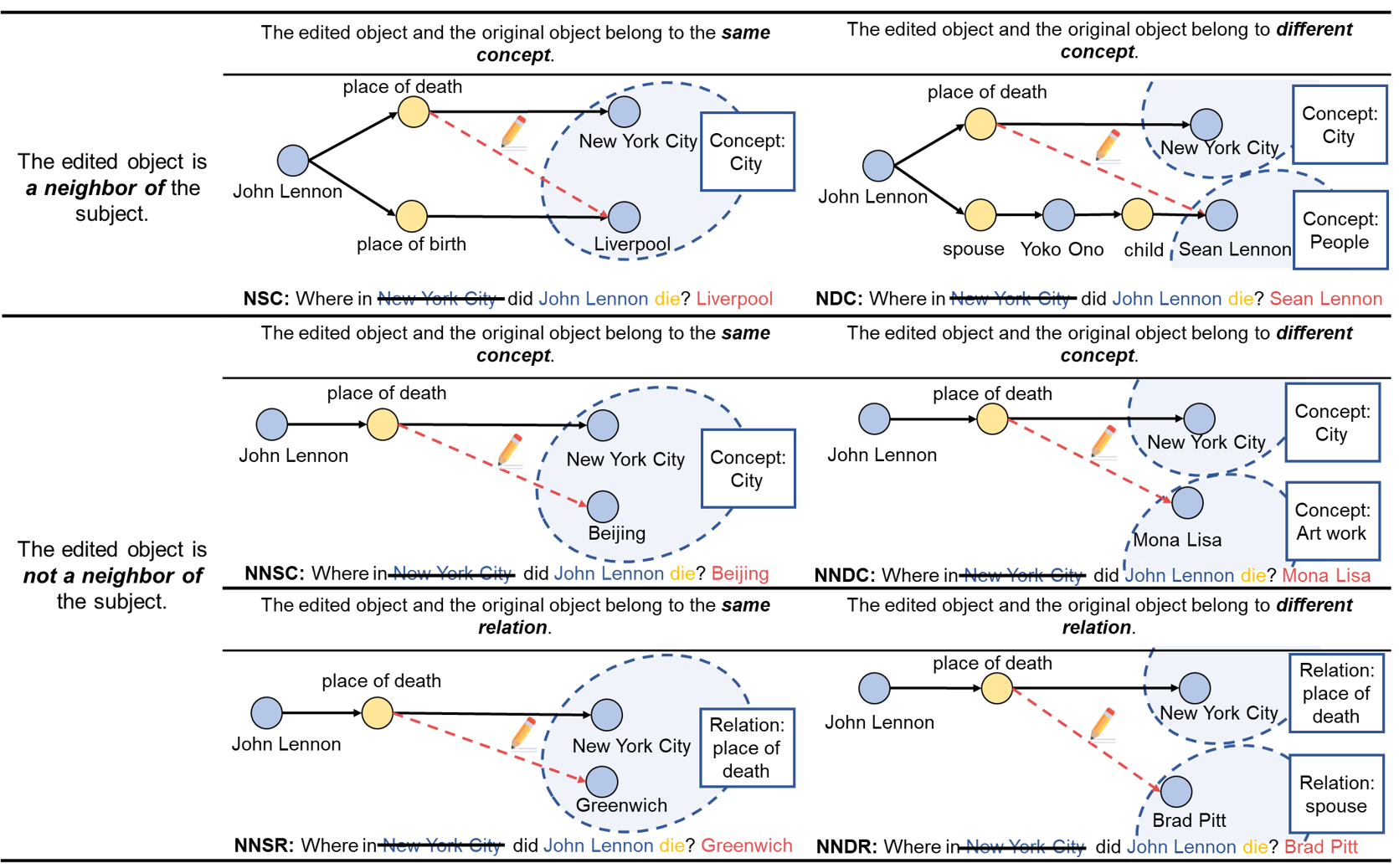}
  \centering
  \caption{An illustration of editing methods in KG-FPQ. We use acronyms to refer each method: Neighbor-Same-Concept (NSC), Neighbor-Different-Concept (NDC), Not-Neighbor-Same-Concept (NNSC), Not-Neighbor-Different-Concept (NNDC), Not-Neighbor-Same-Relation (NNSR), Not-Neighbor-Different-Relation (NNDR).} 
  \label{fig:edting}
\end{figure*}

\noindent\textbf{False Premise Questions}
Existing FPQ benchmarks \citep{yu-etal-2023-crepe, kim-etal-2023-qa, hu-etal-2023-wont, vu2023freshllms} primarily rely on manual construction, resulting in limited scale, lack of extensibility and high labor costs. \citet{yuan2024whispers} construct their dataset by corrupting triplets in Wikidata \citep{10.1145/2629489} and filling them into human-written templates. However, the dataset covers only two narrow topics, and the use of fixed templates limits its semantic richness.  Additionally, these studies lack a thorough evaluation of factuality hallucination induced by FPQs. KG-FPQ is automatically constructed and offers multiple perspectives for evaluation and analysis.

\section{KG-FPQ Benchmark Construction}
% \subsection{Benchmark Construction}
\subsection{Triplets Extraction and Editing\label{sec:3.1}}
We utilize KoPL\footnote{\url{https://github.com/THU-KEG/KoPL}}~\citep{cao2022kqaprodatasetexplicit}, a high-quality subset of Wikidata, as our KG. KoPL contains a limited set of concepts and relations, where each entity uniquely belongs to one concept. We follow the steps shown in the left of Figure~\ref{fig:pipline} to extract and edit triplets. First, we select entities from three domains: Art, People and Place, based on their concepts, and filter the relations for each domain. The filtering rules are detailed in Appendix~\ref{sec:a1}, and Table~\ref{tab: domain exp} lists the representative concepts, relations, and entities for each domain.\par

Subsequently, we extract true triplets from KoPL and edit them into false triplets. The editing methods, illustrated in Figure~\ref{fig:edting}, can be categorized into six types across two perspectives:  1) the edited object at varying distances from the subject in the KG; 2) the edited object having varying associations with the original object in the KG. In detail, when the edited object is a neighbor of the subject, their maximum distance is set to five hops. Through editing, we get six different false triplets for each true triplet, resulting in six corresponding FPQs during data generation. False triplets created by different editing methods exhibit varying levels of confusability. For instance, as shown in Figure~\ref{fig:edting}, Neighbor-Same-Concept (NSC) indicates that the edited object, \textit{Liverpool}, is a 1-hop neighbor of the subject and belongs to the same concept as the original object, which might be challenging for LLMs to recognize. In contrast, Not-Neighbor-Different-Concept (NNDC) indicates that the edited object, \textit{Mona Lisa}, is not a neighbor of the subject and belongs to a different concept from the original object, making it somewhat easier to identify.\par

\subsection{Data Generation and Verification} 
As shown in the right of Figure~\ref{fig:pipline}, firstly, we sample 1k triplets to assess the quality of FPQ data generated using a combination of KG and GPTs. A manual verification of the generation results for the sampled 1k triplets reveals several issues that occurred during the data generation process. Corresponding measures are implemented in subsequent full-scale data generation to address these problems. 
Secondly, we generate the full dataset, utilizing GPT-3.5 to create Yes-No questions and GPT-4 to create WH-questions\footnote{The GPT-3.5 models used in this paper are all GPT-3.5-turbo-1106 version, and the GPT-4 models are all GPT-4-1106-preview version.}. We prompt GPTs to generate TPQs based on true triplets and then replace the original object with the edited object from false triplets through string matching. Therefore, we create one TPQ and six FPQs in each format based on each true triplet, with these FPQs in each format differing only in the edited object. Finally, to ensure data quality, we perform a thorough manual review of the whole dataset, with particular attention to WH-questions, correcting some grammatical and semantic errors. Details on the small-scale data generation process and the manual review procedure are provided in Appendix~\ref{sec:a2}, while the prompt templates utilized for data generation are included in Appendix~\ref{sec:a3}.

\section{Experiment Settings} 
\subsection{Tasks}
\noindent\textbf{Discriminative Task} 
The first task involves the discriminative task, where LLMs are required to answer Yes-No questions in KG-FPQ with ``Yes'' or ``No'' only, without providing explanations. An example for FPQ in Yes-No format is that \textit{Did John Lennon die in Liverpool?}.
\\

\noindent\textbf{Generative Task} 
The second task involves the generative task, where LLMs are required to answer the WH-questions in KG-FPQ. An example for FPQ in WH format is that \textit{Where in Liverpool did John Lennon die?}. If LLMs recognize the false premises in FPQs, they will deny the false premises and provide explanations. If LLMs fail to identify the false premises, they may be misled by FPQs and generate information with fctuality hallucination.

\begin{figure}[t] 
  \includegraphics[width=\linewidth]{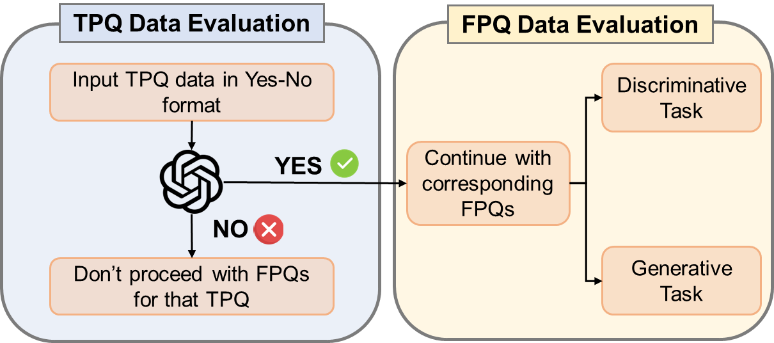}
  \centering
  \caption{Overview of the evaluation procedure.} 
  \label{fig:evaluation procedure}
\end{figure}

\subsection{Models\label{sec:models}}
We select several representative and advanced open-source chat models of various sizes. Models in the 6B\textasciitilde8B range include ChatGLM3-6B~\citep{du2022glm}, Baichuan2-7B-Chat~\citep{baichuan2023baichuan2}, Llama2-7B-Chat~\citep{touvron2023llama}, Qwen1.5-7B-Chat~\citep{qwen}, and Llama3-8B-instruction~\citep{metaai2024llama3herdmodels}. Models in the 13B\textasciitilde14B range include Baichuan2-13B-Chat~\citep{baichuan2023baichuan2}, Llama2-13B-Chat~\citep{touvron2023llama}, and Qwen1.5-14B-Chat~\citep{qwen}. We also evaluate advanced two closed-source LLMs, GPT-3.5~\citep{openai2023gpt35} and GPT-4~\citep{openai2024gpt4} on the discriminative task. We set the temperature parameter to 0.6 and the top\_p parameter to 0.9 for all models in both the discriminative task and the generative task.

\subsection{Evaluation\label{sec:evaluation}}
\noindent\textbf{Evaluation Procedure}
Our evaluation procedure is shown in Figure~\ref{fig:evaluation procedure}. First, we input the Yes-No format TPQs into the LLMs. If the LLMs answer ``Yes'', they are considered correct, which indicates that the LLMs have stored relevant background knowledge for the question. We then continue with the corresponding FPQs in both the discriminative task and the generative task. If the LLMs answer ``No'' to a TPQ, we do not proceed with the FPQs for that TPQ. This approach aims to reduce the hallucination caused by a lack of background knowledge. To increase the robustness of the assessment, we input each question three times to obtain three responses, and then perform a hard vote to get the final answer label. The prompt templates for evaluation are presented in Appendix~\ref{sec:b1}.
\\

\noindent\textbf{Evaluation for Generative Task}
Since manual evaluation of the generative task is costly, we introduce an automated evaluator named FPQ-Judge, which is a LoRA-tuned Llama3-8B-instruction model designed to classify whether the answers of LLMs to FPQs are misled by the false premises. The training set for FPQ-Judge consists of triplets in the form of (question, answer, label), where the label indicates whether the answer is true or false. This training set includes 13k examples where the answer is a true/false reference answer generated by GPT-3.5. Additionally, it comprises approximately 15k examples where the answer is generated by one of the evaluated models in Section~\ref{sec:models}, with the label derived from human annotation. To assess the performance of FPQ-Judge, we conduct tests on both a GPT-3.5 generated test set with a size of 3k and a human annotated test set with a size of 6.3k. FPQ-Judge achieves an accuracy of 99.32\% on the GPT-3.5 generated test set and 93\% on the manually annotated test set. The prompt templates used for GPT-3.5 to generate traing data, the examples of the training data, and the training parameters are provided in the Appendix~\ref{sec:b2}.
\\

\noindent\textbf{Metrics}
We use accuracy as the evaluation metric. In the discriminative task, we calculate accuracy by string matching the responses of LLMs: for TPQs, answering ``Yes'' is considered correct; for FPQs, answering ``No'' is considered correct. In the generative task, an answer is considered correct if FPQ-Judge marks it as correct~\footnote{FPQ-Judge can't ensure the answer is completely non-hallucinated.}.

\begin{table*}
    \centering
    \small
    \begin{tabular}{lm{0.6cm}m{0.6cm}m{0.6cm}m{0.6cm}m{0.6cm}m{0.6cm}m{0.6cm}m{0.6cm}m{0.6cm}m{0.6cm}m{0.6cm}m{0.6cm}}
    \toprule
        \multirow{2}{*}{\textbf{Model}} & \multicolumn{6}{c}{\textbf{Art Dis}} & \multicolumn{6}{c}{\textbf{Art Gen}}  \\
        \cmidrule(lr){2-7} \cmidrule(lr){8-13}
        & \textbf{NSC} & \textbf{NDC} & \textbf{NNSC} & \textbf{NNDC} & \textbf{NNSR} & \textbf{NNDR} & \textbf{NSC} & \textbf{NDC} & \textbf{NNSC} & \textbf{NNDC} & \textbf{NNSR} & \textbf{NNDR}  \\
        \midrule
        ChatGLM3-6B & 0.561 & 0.797 & 0.644 & 0.836 & 0.572 & 0.805 & 0.215 & 0.224 & 0.189 & 0.231 & 0.168 & 0.237\\
        Baichuan2-7B-Chat & 0.412 & 0.571 & 0.507 & 0.634 & 0.423 & 0.61 & 0.454 & 0.461 & 0.493 & 0.534 & 0.42 & 0.539\\
        Qwen1.5-7B-Chat & 0.742 & 0.903 & 0.835 & 0.952 & 0.803 & 0.948 & 0.503 & 0.586 & 0.606 & 0.673 & 0.526 &	0.682\\
        Llama2-7B-Chat&	0.722 & 0.81 & 0.792 & 0.857 & 0.783 & 0.845 & 0.446 & 0.429 &	0.488 & 0.513 &	0.463 &	0.494\\
        Llama3-8B-instruct & 0.77 & 0.9 & 0.891 & 0.959 & 0.868 & 0.951 & 0.644 &	0.556 &	0.725 &	0.664 &	0.707 &	0.68\\
        \midrule
        Baichuan2-13B-Chat	& 0.414 & 0.588 & 0.484 & 0.669 & 0.409 & 0.652 & 0.309 & 0.269 &	0.336 &	0.324 &	0.303 &	0.341\\
        Qwen1.5-14B-Chat & 0.806 & 0.941 & 0.893 & 0.989 & 0.857 & 0.986 &
        0.389 & 0.445 &	0.469 &	0.528 &	0.409 &	0.539\\
        Llama2-13B-Chat & 0.876 & 0.95 & 0.956 &0.988 & 0.962 & 0.982 & 0.879	& 0.867 &	0.926 & 0.924 & 0.921 &	0.923\\
        \midrule
        GPT-3.5 & 0.808 & 0.862	& 0.829	& 0.92	& 0.741	& 0.898 & - & - & - & - & - & -\\
        GPT-4 & 0.874 &	0.963 &	0.977 &	0.988 & 0.96 &	0.994 & - & - & - & - & - & -\\
        \midrule
        average acc & 0.698 & 0.829 & 0.781 & 0.879 & 0.738 & 0.867& 0.48 &	0.482&	0.529&	0.549&	0.49&	0.55\\
    \bottomrule
    \end{tabular}
  \caption{The evaluation results for FPQs on the discriminative task (referred to as Dis) and the generative task (referred to as Gen) in Art domain.}
  \label{tab: Art result}
\end{table*}

\section{Results} 
Table~\ref{tab:fpq} presents the complete evaluation results of all models for FPQs on both the discriminative task and the generative task across three domains. Table~\ref{tab: Art result} presents the results of the Art domain, which we use as an example for preliminary analysis. It can be observed that the accuracy of LLMs varies across FPQs with different levels of confusability, and their performance also differs based on the task format. In Section 5.1, we will further analyze the relationship between the confusability of FPQs and the factuality hallucination. In Section 5.2, we will examine the impact of task format on factuality hallucination. Additionally, Section 5.3 and Section 5.4 will provide detailed analyses from the perspectives of knowledge domains and model scales, respectively. 

\subsection{Impact of confusability of FPQs\label{sec:5.1}}
As shown in Figure~\ref{fig:edting}, we design editing methods from two perspectives, \textit{distance} and \textit{association}, and create FPQs at six levels of confusability. In this section we will discuss the impact of confusability of FPQs from these two perspectives.

\subsubsection{Impact of Distance\label{sec:distance}}
\begin{figure}[t] 
  \includegraphics[width=\linewidth]{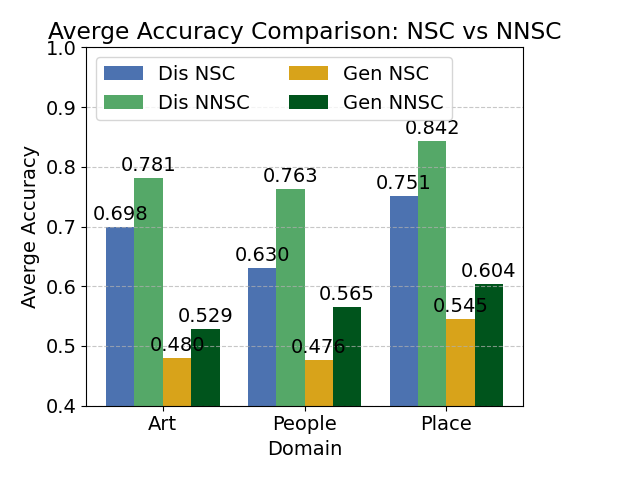}
  \centering
  \caption{The average accuracy of all models comparison between NSC and NNSC.} 
  \label{fig:NSCvsNNSC}
\end{figure}

To investigate the impact of the distance between the edited object and the subject within the KG, the average accuracy of all LLMs on NSC and NNSC is calculated in both discriminative and generative tasks across three domains, as illustrated in Figure~\ref{fig:NSCvsNNSC}. The results demonstrate that, the average accuracy for NSC is consistently lower than for NNSC across all domains, and this phenomenon is more pronounced in the discriminative task. This indicates that FPQs formed when the edited object in the false triplets is a neighbor of the subject are more confusing to LLMs, resulting in a higher probability of factuality hallucination.\par

\begin{figure*}[t] 
  \includegraphics[width=\linewidth]{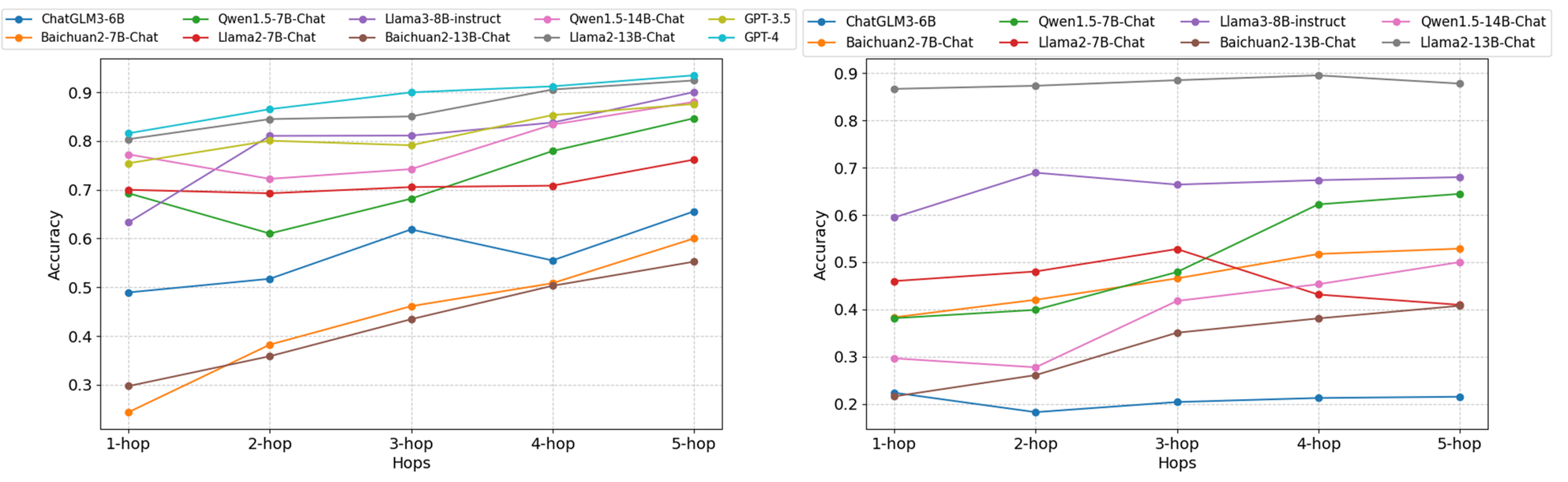}
  \centering
  \caption{Accuracy of all models for NSC in Art domain by hops. Left: Results of the discriminative task. Right: Results of the generative task.} 
  \label{fig:art_hops}
\end{figure*}

Furthermore, we conduct a more detailed examination of NSC and NDC to investigate the impact of the number of hops between the edited object and the subject. The complete results are shown in Appendix~\ref{sec:c1}, and we analyze the NSC in Art domain as an example in this section, with results presented in Figure~\ref{fig:art_hops}. It is observed that for most models, the accuracy improves as the number of hops between the edited object and the subject increases, indicating a reduction in factuality hallucination, and this trend is more evident in discriminative tasks.\par
In conclusion, \textbf{when the edited object and the subject in the false triplets has a closer distance, the FPQs are more confusing for LLMs, and more likely to cause factuality hallucination}. Conversely, as the distance between them increases, the likelihood of factuality hallucination decreases. This trend is more pronounced in the discriminative task than in the generative task.

\subsubsection{Impact of Associations \label{sec:association}}
\begin{figure}[t]
  % \captionsetup{width=0.5\textwidth}  % 调整标题的宽度
  \centering 
  \subfigure{
  \includegraphics[width=0.46\linewidth]{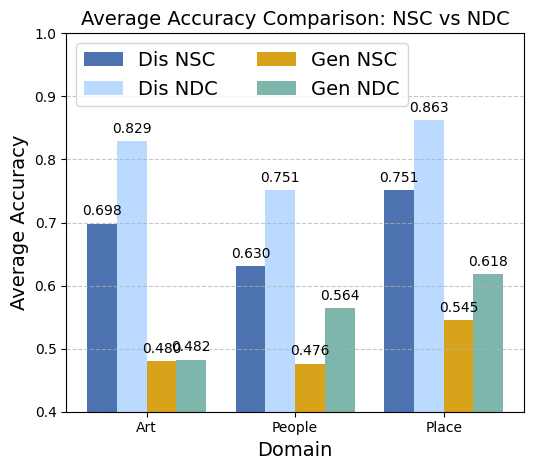}
  }\hfill
   \subfigure{
  \includegraphics[width=0.46\linewidth]{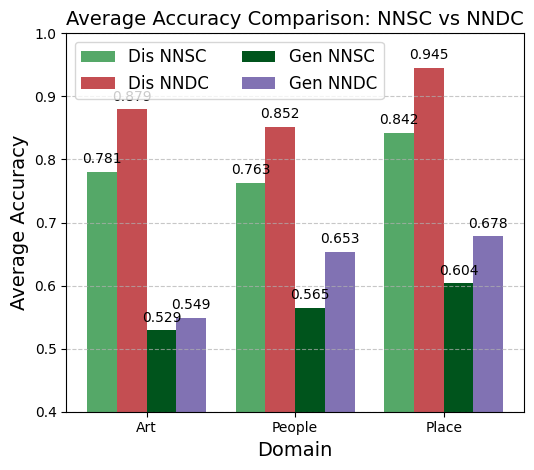}
  }\hfill
  \subfigure{
  \includegraphics[width=0.46\linewidth]{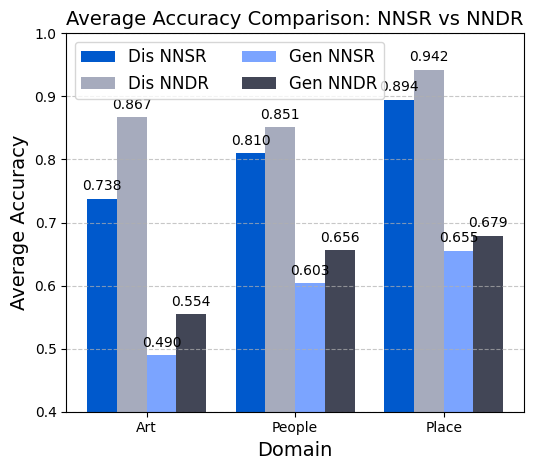}
  }\hfill
  \subfigure{
  \includegraphics[width=0.46\linewidth]{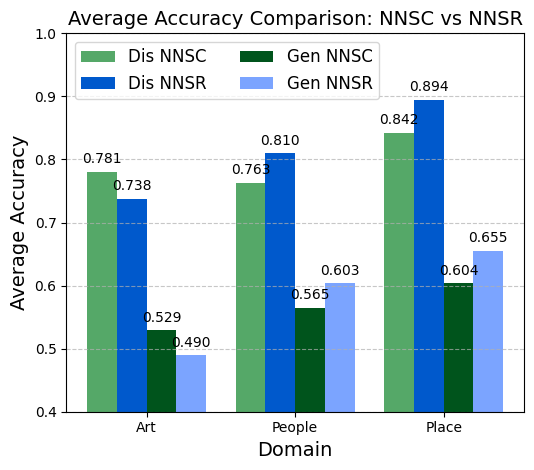}
  }
  \caption{The average accuracy comparison. Upper Left: NSC vs. NDC. Upper Right: NNSC vs. NNDC. Lower Left: NNSR vs. NNDR. Lower Right: NNSC vs. NNSR.}
 \label{fig:editing method}
 \vspace{-5mm}
\end{figure}

To explore the impact of the associations between the edited object and the original object on FPQs-induced factuality hallucination, we calculate the average accuracy of all LLMs on NSC vs. NDC, NNSC vs. NNDC, NNSR vs. NNDR, and NNSC vs. NNSR in both tasks across three domains, as illustrated in Figure~\ref{fig:editing method}.\par

From the comparison of NSC vs. NDC, and NNSC vs. NNDC in upper Figure~\ref{fig:editing method}, it is evident that in all domains, whether in the discriminative or generative task, the average accuracy for NSC is consistently lower than for NDC, and as the same, NNSC is consistently lower than NNDC. As shown in Figure~\ref{fig:edting}, NSC and NNSC involve the edited object and original object belonging to the ~\textit{same concept} in the KG, whereas NDC and NNDC involve ~\textit{different concepts}. Thus, we conclude that when the edited object and the original object belong to the \textit{same concept} in the KG, the FPQs generated are more confusing for LLMs, leading to a higher likelihood of factuality hallucination. Similarly, the comparison between NNSR and NNDR in the lower left of Figure~\ref{fig:editing method} reveals that FPQs generated from false triplets where the edited object and original object share the ~\textit{same relation} are more likely to induce factuality hallucination in LLMs.\par
We also compare NNSC and NNSR to determine whether the ~\textit{same concept} or the ~\textit{same relation} editing method has a greater impact on LLMs. The lower right of Figure~\ref{fig:editing method} shows that in the Art domain, the NNSC creates stronger interference than the NNSR, while in the People and Place domains, the NNSR causes greater interference.\par
In conclusion, \textbf{when the edited object has stronger associations with the original object, the FPQs are more confusing for LLMs, and likely to induce factuality hallucination}.

\subsection{Impact of Task Format\label{sec:task format}}
\begin{figure}[t] 
  \includegraphics[width=\linewidth]{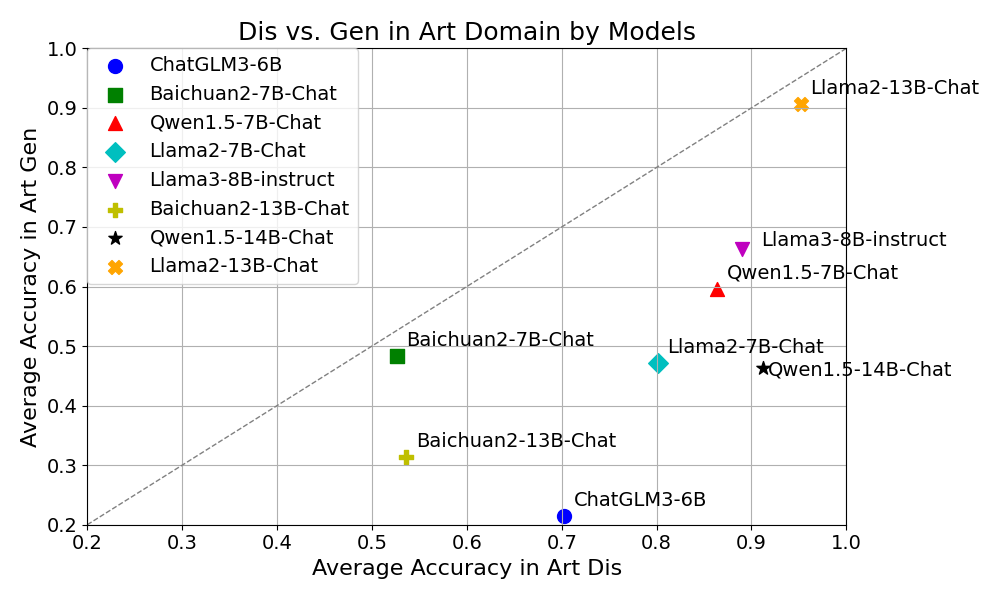}
  \centering
  \caption{The overall performance comparison between the discriminative task and the generative task by models in Art domain.} 
  \label{fig:disvsgen}
\end{figure}

We analyze the overall performance of each LLM in both discriminative and generative tasks, with the complete results shown in Appendix~\ref{sec:c2}. This section provides an analysis of the Art domain, with results presented in Figure~\ref{fig:disvsgen}. It is evident that for almost all LLMs, the overall accuracy in the generative task is lower than in the discriminative task, suggesting that \textbf{LLMs perform worse at generating factual statements than at distinguishing them when faced with FPQs}. This highlights that generative FPQs remain a significant challenge for LLMs and warrant further attention. 

\subsection{Impact of Knowledge Domain}
\begin{figure}[t] 
  \includegraphics[width=\linewidth]{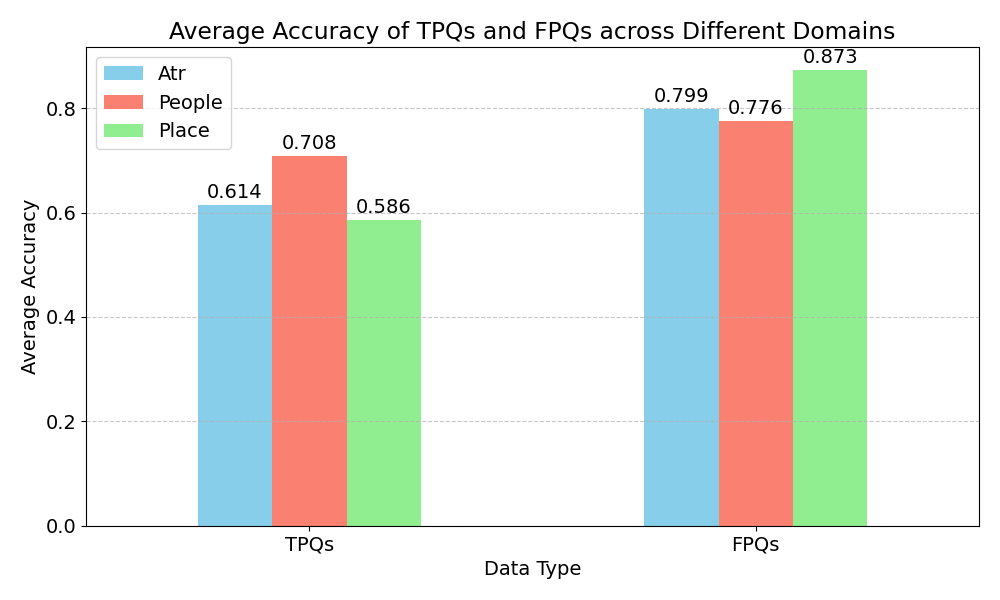}
  \centering
  \caption{The average accuracy of TPQs and FPQs across domains.} 
  \label{fig:tpqvsfpq in domain}
\end{figure}

Following the procedure shown in Figure~\ref{fig:evaluation procedure}, we first evaluate LLMs on Yes-No format TPQs, with the results presented in Appendix~\ref{sec:c3}. We propose a hypothesis: From the domain perspective, higher accuracy on TPQs indicates that LLMs are more familiar with the knowledge in that domain, and therefore, the accuracy on FPQs in that domain should also be higher, implying that LLMs are less likely to be misled by FPQs. To verify it, we compare the results of TPQs and FPQs, shown in Figure~\ref{fig:tpqvsfpq in domain}. The average accuracy of TPQs is higher in the People domain compared to Art and Place, whereas the average accuracy of FPQs is highest in the Place domain compared to Art and People. This indicates that \textbf{the knowledge proficiency of LLMs varies across domains, and that there is no positive correlation between knowledge proficiency and the ability to resist the interference of FPQs}.

\subsection{Impact of Model size\label{sec:model size}}
\begin{figure}[t]
  % \captionsetup{width=0.5\textwidth}  % 调整标题的宽度
\includegraphics[width=\linewidth]{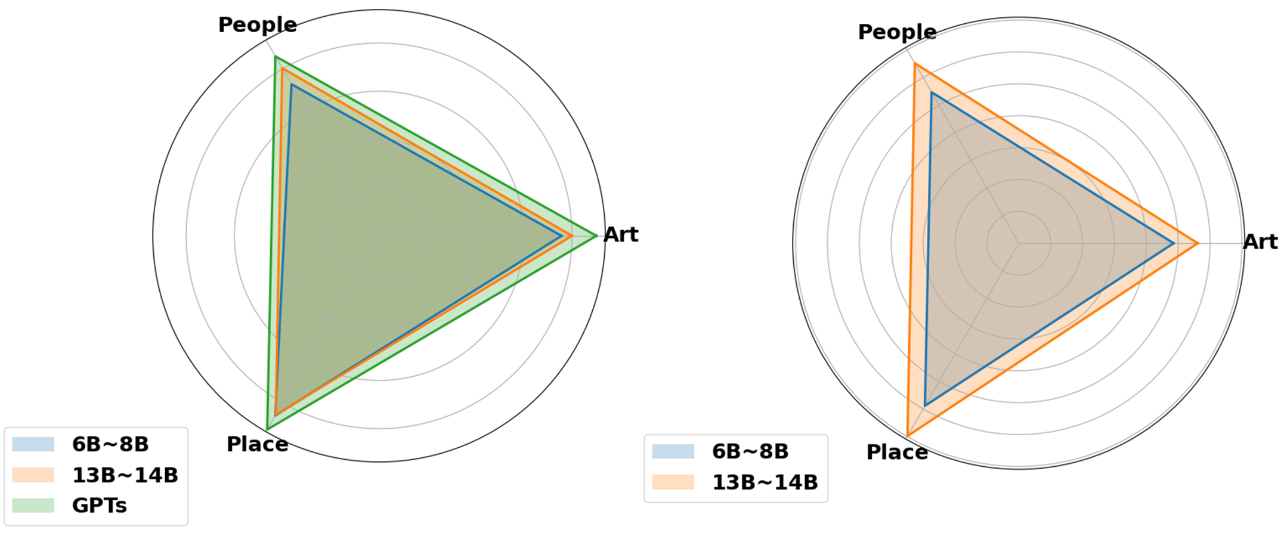}
  \centering
  \caption{The average accuracy of FPQs comparison across different model size. Left: Results for the discriminative task. Right: Results for the generative task.}
 \label{fig:model size}
\end{figure}

The evaluated models are classified into 3 categories according to their size: 6B\textasciitilde8B, 13B\textasciitilde14B, and the GPT series. We then calculate the average accuracy for FPQs of each categories across 3 domains, as shown in Figure~\ref{fig:model size}. It can be observed that, regardless of the task format, the average accuracy tends to increase with larger model sizes. This indicates that \textbf{larger models are more factual in answering FPQs}. A similar analysis is conducted for TPQs as presented in Appendix~\ref{sec:c4}. The GPT series demonstrate the highest performance on TPQs, while the 6B\textasciitilde8B LLMs outperform the 13B\textasciitilde14B LLMs, which is counterintuitive.\par

Observing Table~\ref{tab:tpq}, we find that the accuracy of the Baichuan2 series is significantly higher than that of other models, and the accuracy of Llama2-13B-Chat is even far below the random guessing probability of 0.5. We undertake a closer examination of these three models, and the results are shown in Figure~\ref{fig:models}. In most cases, the performance of FPQs for the Baichuan2 series decreases compared to TPQs. By contrast, the accuracy of FPQs for Llama2-13B-Chat significantly increases compared to TPQs. We hypothesize that these models may have an inherent bias that causes them to consistently favor one type of answer when answering Yes-No questions. Despite using repeated questioning and hard voting strategies during evaluation, this tendency remains noticeable, which should be addressed by developers.

\section{Conclusion and Discussion}
To evaluate factual hallucination induced by false premise questions in LLMs, we develop an automated and scalable pipeline to construct FPQs by editing the triplets in a KG and utilizing GPTs to generate data. Based on the proposed method we create a comprehensive benchmark, KG-FPQ, offering multiple perspectives for evaluation. Using KG-FPQ, we assess several advanced LLMs. Through extensive experiments, we reach three essential conclusions: (1) FPQs with different levels of confusability have varying degrees of impact on LLMs. (2) LLMs perform worse at generating factual statements than at distinguishing them when faced with FPQs. (3) Knowledge proficiency of LLMs varies across domains, and there is no positive correlation between knowledge proficiency and the ability to resist the interference of FPQs.\par
Based on analysis in Section~\ref{sec:5.1}, we speculate that the internal knowledge storage structures of LLMs may resemble knowledge graphs, which we will explore further in future research. Additionally, FPQs can be exploited as prompt injection attacks, leading LLMs to generate non-factual texts and spread misinformation online. In order to identify and mitigate more potential vulnerabilities, we will expand the variety of FPQs for red teaming LLMs.

\section*{Limitations}
We propose a comprehensive FPQ benchmark, based on which we evaluate the FPQ-induced factual hallucinations in several advanced LLMs in both discriminative and generative formats. However, our work still faces limitations and challenges. Firstly, the structured knowledge stored in knowledge graphs is difficult to update in line with developments in the real world, which may lead to misjudgments in some cases. Secondly, as mentioned in Section~\ref{sec:model size}, certain models exhibit an inherent bias in the discriminative evaluation, consistently favoring one type of answer when responding to discriminative questions. Although we have taken measures to enhance the robustness of our evaluation, this bias remains unavoidable. Lastly, we fine-tune an evaluator for generative hallucination evaluation, achieving high accuracy in our task. However, this evaluator cannot detect all hallucination in the responses of LLMs, and its generalization performance to other tasks remains to be explored. More precise and comprehensive hallucination detection is still a challenge in the era of LLMs, which we aim to further explore in the future.

\section*{Acknowledgments}
This work is supported by the National Key R\&D Program of China (No. 2023YFC3310700). We sincerely appreciate the anonymous reviewers and area chairs for their invaluable and constructive feedback, which has been instrumental in enhancing the quality of our study.

\bibliography{custom}

\begin{thebibliography}{24}
\providecommand{\natexlab}[1]{#1}

\bibitem[{Bai et~al.(2023)Bai, Bai, Chu, Cui, Dang, Deng, Fan, Ge, Han, Huang, Hui, Ji, Li, Lin, Lin, Liu, Liu, Lu, Lu, Ma, Men, Ren, Ren, Tan, Tan, Tu, Wang, Wang, Wang, Wu, Xu, Xu, Yang, Yang, Yang, Yang, Yao, Yu, Yuan, Yuan, Zhang, Zhang, Zhang, Zhang, Zhou, Zhou, Zhou, and Zhu}]{qwen}
Jinze Bai, Shuai Bai, Yunfei Chu, Zeyu Cui, Kai Dang, Xiaodong Deng, Yang Fan, Wenbin Ge, Yu~Han, Fei Huang, Binyuan Hui, Luo Ji, Mei Li, Junyang Lin, Runji Lin, Dayiheng Liu, Gao Liu, Chengqiang Lu, Keming Lu, Jianxin Ma, Rui Men, Xingzhang Ren, Xuancheng Ren, Chuanqi Tan, Sinan Tan, Jianhong Tu, Peng Wang, Shijie Wang, Wei Wang, Shengguang Wu, Benfeng Xu, Jin Xu, An~Yang, Hao Yang, Jian Yang, Shusheng Yang, Yang Yao, Bowen Yu, Hongyi Yuan, Zheng Yuan, Jianwei Zhang, Xingxuan Zhang, Yichang Zhang, Zhenru Zhang, Chang Zhou, Jingren Zhou, Xiaohuan Zhou, and Tianhang Zhu. 2023.
\newblock Qwen technical report.
\newblock \emph{arXiv preprint arXiv:2309.16609}.

\bibitem[{Baichuan(2023)}]{baichuan2023baichuan2}
Baichuan. 2023.
\newblock \href {https://arxiv.org/abs/2309.10305} {Baichuan 2: Open large-scale language models}.
\newblock \emph{arXiv preprint arXiv:2309.10305}.

\bibitem[{Cao et~al.(2022)Cao, Shi, Pan, Nie, Xiang, Hou, Li, He, and Zhang}]{cao2022kqaprodatasetexplicit}
Shulin Cao, Jiaxin Shi, Liangming Pan, Lunyiu Nie, Yutong Xiang, Lei Hou, Juanzi Li, Bin He, and Hanwang Zhang. 2022.
\newblock \href {https://arxiv.org/abs/2007.03875} {Kqa pro: A dataset with explicit compositional programs for complex question answering over knowledge base}.
\newblock \emph{Preprint}, arXiv:2007.03875.

\bibitem[{Du et~al.(2022)Du, Qian, Liu, Ding, Qiu, Yang, and Tang}]{du2022glm}
Zhengxiao Du, Yujie Qian, Xiao Liu, Ming Ding, Jiezhong Qiu, Zhilin Yang, and Jie Tang. 2022.
\newblock Glm: General language model pretraining with autoregressive blank infilling.
\newblock In \emph{Proceedings of the 60th Annual Meeting of the Association for Computational Linguistics (Volume 1: Long Papers)}, pages 320--335.

\bibitem[{Hu et~al.(2023)Hu, Luo, Wang, Cheng, Liu, and Sun}]{hu-etal-2023-wont}
Shengding Hu, Yifan Luo, Huadong Wang, Xingyi Cheng, Zhiyuan Liu, and Maosong Sun. 2023.
\newblock \href {https://doi.org/10.18653/v1/2023.acl-long.309} {Won{'}t get fooled again: Answering questions with false premises}.
\newblock In \emph{Proceedings of the 61st Annual Meeting of the Association for Computational Linguistics (Volume 1: Long Papers)}, pages 5626--5643, Toronto, Canada. Association for Computational Linguistics.

\bibitem[{Huang et~al.(2023)Huang, Yu, Ma, Zhong, Feng, Wang, Chen, Peng, Feng, Qin, and Liu}]{huang2023survey}
Lei Huang, Weijiang Yu, Weitao Ma, Weihong Zhong, Zhangyin Feng, Haotian Wang, Qianglong Chen, Weihua Peng, Xiaocheng Feng, Bing Qin, and Ting Liu. 2023.
\newblock \href {https://arxiv.org/abs/2311.05232} {A survey on hallucination in large language models: Principles, taxonomy, challenges, and open questions}.
\newblock \emph{Preprint}, arXiv:2311.05232.

\bibitem[{Kim et~al.(2023)Kim, Htut, Bowman, and Petty}]{kim-etal-2023-qa}
Najoung Kim, Phu~Mon Htut, Samuel~R. Bowman, and Jackson Petty. 2023.
\newblock \href {https://doi.org/10.18653/v1/2023.acl-long.472} {({QA})$^2$: Question answering with questionable assumptions}.
\newblock In \emph{Proceedings of the 61st Annual Meeting of the Association for Computational Linguistics (Volume 1: Long Papers)}, pages 8466--8487, Toronto, Canada. Association for Computational Linguistics.

\bibitem[{Li et~al.(2023)Li, Cheng, Zhao, Nie, and Wen}]{li2023halueval}
Junyi Li, Xiaoxue Cheng, Wayne~Xin Zhao, Jian-Yun Nie, and Ji-Rong Wen. 2023.
\newblock \href {https://arxiv.org/abs/2305.11747} {Halueval: A large-scale hallucination evaluation benchmark for large language models}.
\newblock \emph{Preprint}, arXiv:2305.11747.

\bibitem[{Lin et~al.(2022)Lin, Hilton, and Evans}]{lin2022truthfulqa}
Stephanie Lin, Jacob Hilton, and Owain Evans. 2022.
\newblock \href {https://arxiv.org/abs/2109.07958} {Truthfulqa: Measuring how models mimic human falsehoods}.
\newblock \emph{Preprint}, arXiv:2109.07958.

\bibitem[{Liu et~al.(2019)Liu, Ott, Goyal, Du, Joshi, Chen, Levy, Lewis, Zettlemoyer, and Stoyanov}]{liu2019robertarobustlyoptimizedbert}
Yinhan Liu, Myle Ott, Naman Goyal, Jingfei Du, Mandar Joshi, Danqi Chen, Omer Levy, Mike Lewis, Luke Zettlemoyer, and Veselin Stoyanov. 2019.
\newblock \href {https://arxiv.org/abs/1907.11692} {Roberta: A robustly optimized bert pretraining approach}.
\newblock \emph{Preprint}, arXiv:1907.11692.

\bibitem[{Meta(2024)}]{metaai2024llama3herdmodels}
Meta. 2024.
\newblock Introducing meta llama 3: The most capable openly available llm to date.
\newblock \url{https://ai.meta.com/blog/meta-llama-3/}.

\bibitem[{Min et~al.(2023)Min, Krishna, Lyu, Lewis, tau Yih, Koh, Iyyer, Zettlemoyer, and Hajishirzi}]{min2023factscore}
Sewon Min, Kalpesh Krishna, Xinxi Lyu, Mike Lewis, Wen tau Yih, Pang~Wei Koh, Mohit Iyyer, Luke Zettlemoyer, and Hannaneh Hajishirzi. 2023.
\newblock \href {https://arxiv.org/abs/2305.14251} {Factscore: Fine-grained atomic evaluation of factual precision in long form text generation}.
\newblock \emph{Preprint}, arXiv:2305.14251.

\bibitem[{Muhlgay et~al.(2024)Muhlgay, Ram, Magar, Levine, Ratner, Belinkov, Abend, Leyton-Brown, Shashua, and Shoham}]{muhlgay2024generating}
Dor Muhlgay, Ori Ram, Inbal Magar, Yoav Levine, Nir Ratner, Yonatan Belinkov, Omri Abend, Kevin Leyton-Brown, Amnon Shashua, and Yoav Shoham. 2024.
\newblock \href {https://arxiv.org/abs/2307.06908} {Generating benchmarks for factuality evaluation of language models}.
\newblock \emph{Preprint}, arXiv:2307.06908.

\bibitem[{OpenAI(2023)}]{openai2023gpt35}
OpenAI. 2023.
\newblock Gpt-3.5.
\newblock \url{https://www.openai.com/gpt-3.5}.

\bibitem[{OpenAI(2024{\natexlab{a}})}]{chatgpt2024}
OpenAI. 2024{\natexlab{a}}.
\newblock Chatgpt.
\newblock \url{https://chat.openai.com}.

\bibitem[{OpenAI(2024{\natexlab{b}})}]{openai2024gpt4}
OpenAI. 2024{\natexlab{b}}.
\newblock Gpt-4.
\newblock \url{https://www.openai.com/gpt-4}.

\bibitem[{Raffel et~al.(2023)Raffel, Shazeer, Roberts, Lee, Narang, Matena, Zhou, Li, and Liu}]{raffel2023exploringlimitstransferlearning}
Colin Raffel, Noam Shazeer, Adam Roberts, Katherine Lee, Sharan Narang, Michael Matena, Yanqi Zhou, Wei Li, and Peter~J. Liu. 2023.
\newblock \href {https://arxiv.org/abs/1910.10683} {Exploring the limits of transfer learning with a unified text-to-text transformer}.
\newblock \emph{Preprint}, arXiv:1910.10683.

\bibitem[{Touvron et~al.(2023)Touvron, Martin, Stone, Albert, Almahairi, Babaei, Bashlykov, Batra, Bhargava, Bhosale, Bikel, Blecher, Ferrer, Chen, Cucurull, Esiobu, Fernandes, Fu, Fu, Fuller, Gao, Goswami, Goyal, Hartshorn, Hosseini, Hou, Inan, Kardas, Kerkez, Khabsa, Kloumann, Korenev, Koura, Lachaux, Lavril, Lee, Liskovich, Lu, Mao, Martinet, Mihaylov, Mishra, Molybog, Nie, Poulton, Reizenstein, Rungta, Saladi, Schelten, Silva, Smith, Subramanian, Tan, Tang, Taylor, Williams, Kuan, Xu, Yan, Zarov, Zhang, Fan, Kambadur, Narang, Rodriguez, Stojnic, Edunov, and Scialom}]{touvron2023llama}
Hugo Touvron, Louis Martin, Kevin Stone, Peter Albert, Amjad Almahairi, Yasmine Babaei, Nikolay Bashlykov, Soumya Batra, Prajjwal Bhargava, Shruti Bhosale, Dan Bikel, Lukas Blecher, Cristian~Canton Ferrer, Moya Chen, Guillem Cucurull, David Esiobu, Jude Fernandes, Jeremy Fu, Wenyin Fu, Brian Fuller, Cynthia Gao, Vedanuj Goswami, Naman Goyal, Anthony Hartshorn, Saghar Hosseini, Rui Hou, Hakan Inan, Marcin Kardas, Viktor Kerkez, Madian Khabsa, Isabel Kloumann, Artem Korenev, Punit~Singh Koura, Marie-Anne Lachaux, Thibaut Lavril, Jenya Lee, Diana Liskovich, Yinghai Lu, Yuning Mao, Xavier Martinet, Todor Mihaylov, Pushkar Mishra, Igor Molybog, Yixin Nie, Andrew Poulton, Jeremy Reizenstein, Rashi Rungta, Kalyan Saladi, Alan Schelten, Ruan Silva, Eric~Michael Smith, Ranjan Subramanian, Xiaoqing~Ellen Tan, Binh Tang, Ross Taylor, Adina Williams, Jian~Xiang Kuan, Puxin Xu, Zheng Yan, Iliyan Zarov, Yuchen Zhang, Angela Fan, Melanie Kambadur, Sharan Narang, Aurelien Rodriguez, Robert Stojnic, Sergey Edunov, and Thomas
  Scialom. 2023.
\newblock \href {https://arxiv.org/abs/2307.09288} {Llama 2: Open foundation and fine-tuned chat models}.
\newblock \emph{Preprint}, arXiv:2307.09288.

\bibitem[{Vrande\v{c}i\'{c} and Kr\"{o}tzsch(2014)}]{10.1145/2629489}
Denny Vrande\v{c}i\'{c} and Markus Kr\"{o}tzsch. 2014.
\newblock \href {https://doi.org/10.1145/2629489} {Wikidata: a free collaborative knowledgebase}.
\newblock \emph{Commun. ACM}, 57(10):78–85.

\bibitem[{Vu et~al.(2023)Vu, Iyyer, Wang, Constant, Wei, Wei, Tar, Sung, Zhou, Le, and Luong}]{vu2023freshllms}
Tu~Vu, Mohit Iyyer, Xuezhi Wang, Noah Constant, Jerry Wei, Jason Wei, Chris Tar, Yun-Hsuan Sung, Denny Zhou, Quoc Le, and Thang Luong. 2023.
\newblock \href {https://arxiv.org/abs/2310.03214} {Freshllms: Refreshing large language models with search engine augmentation}.
\newblock \emph{Preprint}, arXiv:2310.03214.

\bibitem[{Yu et~al.(2023)Yu, Min, Zettlemoyer, and Hajishirzi}]{yu-etal-2023-crepe}
Xinyan Yu, Sewon Min, Luke Zettlemoyer, and Hannaneh Hajishirzi. 2023.
\newblock \href {https://doi.org/10.18653/v1/2023.acl-long.583} {{CREPE}: Open-domain question answering with false presuppositions}.
\newblock In \emph{Proceedings of the 61st Annual Meeting of the Association for Computational Linguistics (Volume 1: Long Papers)}, pages 10457--10480, Toronto, Canada. Association for Computational Linguistics.

\bibitem[{Yuan et~al.(2024)Yuan, Cao, Jin, Chen, Zeng, Liu, and Zhao}]{yuan2024whispers}
Hongbang Yuan, Pengfei Cao, Zhuoran Jin, Yubo Chen, Daojian Zeng, Kang Liu, and Jun Zhao. 2024.
\newblock \href {https://arxiv.org/abs/2402.19103} {Whispers that shake foundations: Analyzing and mitigating false premise hallucinations in large language models}.
\newblock \emph{Preprint}, arXiv:2402.19103.

\bibitem[{Zhang et~al.(2023)Zhang, Li, Cui, Cai, Liu, Fu, Huang, Zhao, Zhang, Chen, Wang, Luu, Bi, Shi, and Shi}]{zhang2023sirens}
Yue Zhang, Yafu Li, Leyang Cui, Deng Cai, Lemao Liu, Tingchen Fu, Xinting Huang, Enbo Zhao, Yu~Zhang, Yulong Chen, Longyue Wang, Anh~Tuan Luu, Wei Bi, Freda Shi, and Shuming Shi. 2023.
\newblock \href {https://arxiv.org/abs/2309.01219} {Siren's song in the ai ocean: A survey on hallucination in large language models}.
\newblock \emph{Preprint}, arXiv:2309.01219.

\bibitem[{Zhao et~al.(2023)Zhao, Zhou, Li, Tang, Wang, Hou, Min, Zhang, Zhang, Dong, Du, Yang, Chen, Chen, Jiang, Ren, Li, Tang, Liu, Liu, Nie, and Wen}]{zhao2023survey}
Wayne~Xin Zhao, Kun Zhou, Junyi Li, Tianyi Tang, Xiaolei Wang, Yupeng Hou, Yingqian Min, Beichen Zhang, Junjie Zhang, Zican Dong, Yifan Du, Chen Yang, Yushuo Chen, Zhipeng Chen, Jinhao Jiang, Ruiyang Ren, Yifan Li, Xinyu Tang, Zikang Liu, Peiyu Liu, Jian-Yun Nie, and Ji-Rong Wen. 2023.
\newblock \href {https://arxiv.org/abs/2303.18223} {A survey of large language models}.
\newblock \emph{Preprint}, arXiv:2303.18223.

\end{thebibliography}

\clearpage
\appendix
\section{Benchmark\label{sec:Benchmark}}
\subsection{Filter Rules for Concepts and Relations\label{sec:a1}}
 In KoPL, each entity is associated with a unique concept, such as "Lebron James" being linked to the concept of a "basketball player". The KG comprises 794 distinct concepts, which we have categorized into domains based on common knowledge, thus achieving domain-based classification of entities.\par 
There are 363 relations in KoPL, and we apply the following rules to select relations for each domain: 

\begin{enumerate}[label=\roman*.]
\item The relation is associated with corresponding domain. For example, the relation \textit{continent} is associated with the Place domain but not the Art domain. 
\item The relation is informative and does not cause ambiguity. For example, the relation \textit{sex or gender} is informative and exact, but the relation \textit{family} are relatively ambiguous. 
\end{enumerate}
The data selectors are the co-authors. Table~\ref{tab: domain exp} shows the representative concepts, relations and subjects in KG-FPQ.\par

\subsection{Details on Data Generation and Verification\label{sec:a2}}
The small-scale data generation reveals the following issues:

\begin{enumerate}[label=\roman*.]
\item When the original object in the true triplet is the same as or a substring of the subject, string matching would replace the edited object twice. For example:
    \begin{itemize}
    \item True triplet: <Daredevil, present in, Daredevil>
    \item False triplet: <Daredevil, present in, Czechoslovakia>
    \item TPQ: Is Daredevil present in the work Daredevil?
    \item FPQ: Is Czechoslovakia present in the work Czechoslovakia?
    \end{itemize}
    
\item Triplets containing certain relations result in semantically incoherent sentences, as shown in the example in the upper right of Figure~\ref{fig:pipline}: 
    \begin{itemize}
    \item Triplet: <John Lennon, languages spoken, English>
    \item Yes-No: Is John Lennon languages spoken English?
    \item WH: When is John Lennon languages spoken English?
    \end{itemize}

\item Yes-No questions have minimal grammatical issues, whereas WH-questions have issues with the improper use of special interrogative words.
\end{enumerate}

Based on these findings, we implement the following measures for subsequent data generation:
\begin{itemize}
    \item For \romannumeral1., we exclude triplets where the original object is the same as or a substring of the subject.
    \item For \romannumeral2., we further filter the relations identified in Section~\ref{sec:3.1} based on our sampling experiment experience, resulting in the final set of relations presented in Table~\ref{tab: domain exp}.
    \item For  \romannumeral3., we conduct a manual review of all WH-questions, performed by co-authors of this paper, who are master students in NLP. We correct the WH-questions generated by GPT-4 using our grammatical and semantic knowledge.
\end{itemize}

\subsection{Prompt Templates for Data Generation\label{sec:a3}}
Table~\ref{tab: yes-no prompt} presents the prompt template used for GPT-3.5 to generate Yes-No questions, and Table~\ref{tab: wh prompt} is the prompt template used for GPT-4 to generate WH-questions. We prompt GPTs to generate true premise questions based on true triplets and then replace the original object with the edited object from false triplets through string matching. For each domain, we select three representative true triplets and manually craft them into demonstrations. During generation in each domain, these three demonstrations remain fixed. The instruction is indicated by the yellow text, the demonstrations are represented by the pink text, and the query data is descripited by the purple text.

\section{Experiment Settings\label{sec:Experiments}}
\subsection{Prompt Templates for Evaluation\label{sec:b1}} 
Table~\ref{tab: evaluation prompt} presents the prompt templates used for evaluation.\par

\subsection{FPQ-Judge\label{sec:b2}}

\noindent \textbf{Prompt Templates for Training Data Generation} Table~\ref{tab: factual answers prompt} presents the prompt template used for GPT-3.5 to generate factual answers, and Table~\ref{tab: non-factual answers prompt} is the prompt template used to generate non-facutal answers. For each domain, we select three representative true triplets and manually craft them into demonstrations. During generation in each domain, these three demonstrations remain fixed. The instruction is indicated by the yellow text, the demonstrations are represented by the pink text, and the query data is descripited by the purple text. 
\\

\noindent \textbf{An Example for Training Data} Table~\ref{tab: examples of training data} shows the examples of training data. This training set includes 13k examples where the answer is a true/false reference answer generated by GPT-3.5. Additionally, it comprises approximately 15k examples where the answer is generated by one of the evaluated models from Section~\ref{sec:models}, with the label derived from human annotation. The goal of FPQ-judge is to evaluate truth for the questions in KG-FPQ only, without the need to generalize to new questions. Therefore, we include as many questions as possible in the training set.
\\

\noindent \textbf{Parameters for Fine-tuning} During LoRA fine-tuning, the following parameters are used: 
\begin{itemize}
    \item $r = 8$ (LoRA rank)
    \item $\text{lora\_alpha} = 32$ (LoRA scaling factor)
    \item $\text{lora\_dropout} = 0.05$ (dropout rate)
    \item $\text{learning\_rate} =  1e-4 $
\end{itemize}

\section{Additional Results\label{sec:results}}
Table~\ref{tab:fpq} presents the evaluation results of all models for FPQs on Yes-No Question Task and WH-Question Task.
\subsection{Impact fo Distance\label{sec:c1}}
In NSC and NDC, we categorize FPQs into five types based on the number of hops as shown in Table~\ref{tab:hops}, and calculate the accuracy for each category. The formula is as follows:
\[
\text{accuracy} = \frac{\text{correct number in each category}}{\text{total number in each category}}
\]

\begin{figure*}[t] 
  \subfigure{
  \includegraphics[width=\linewidth]{fpq6_art_hop.png}
  }\hfill
  \subfigure{
  \includegraphics[width=\linewidth]{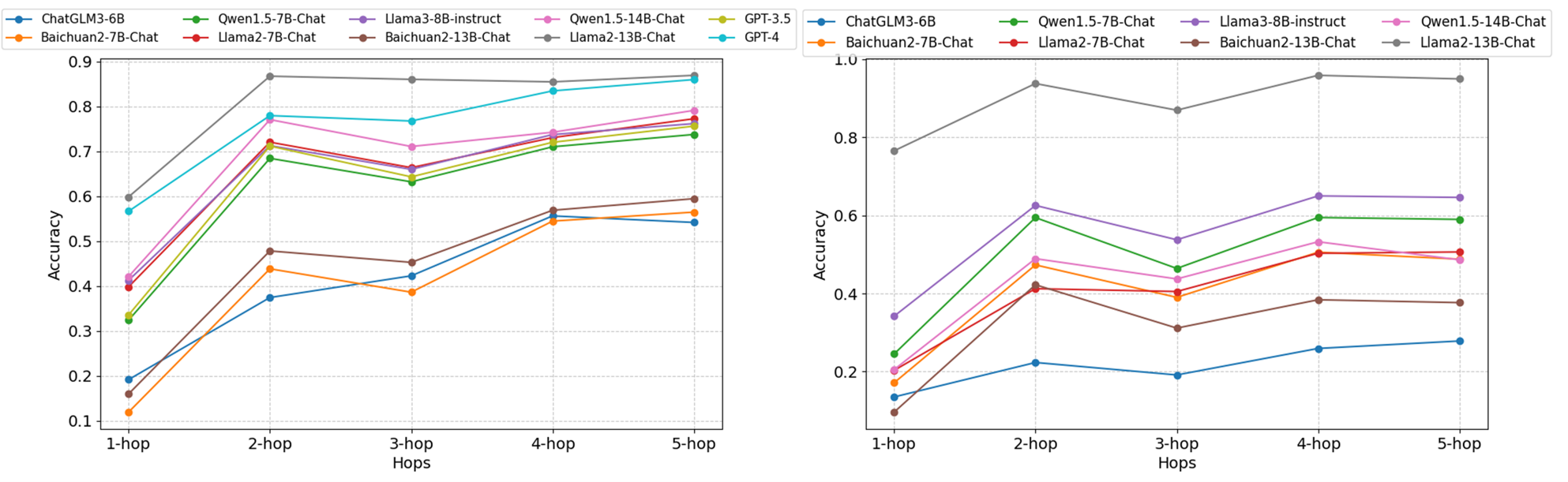}
  }\hfill
  \subfigure{
   \includegraphics[width=\linewidth]{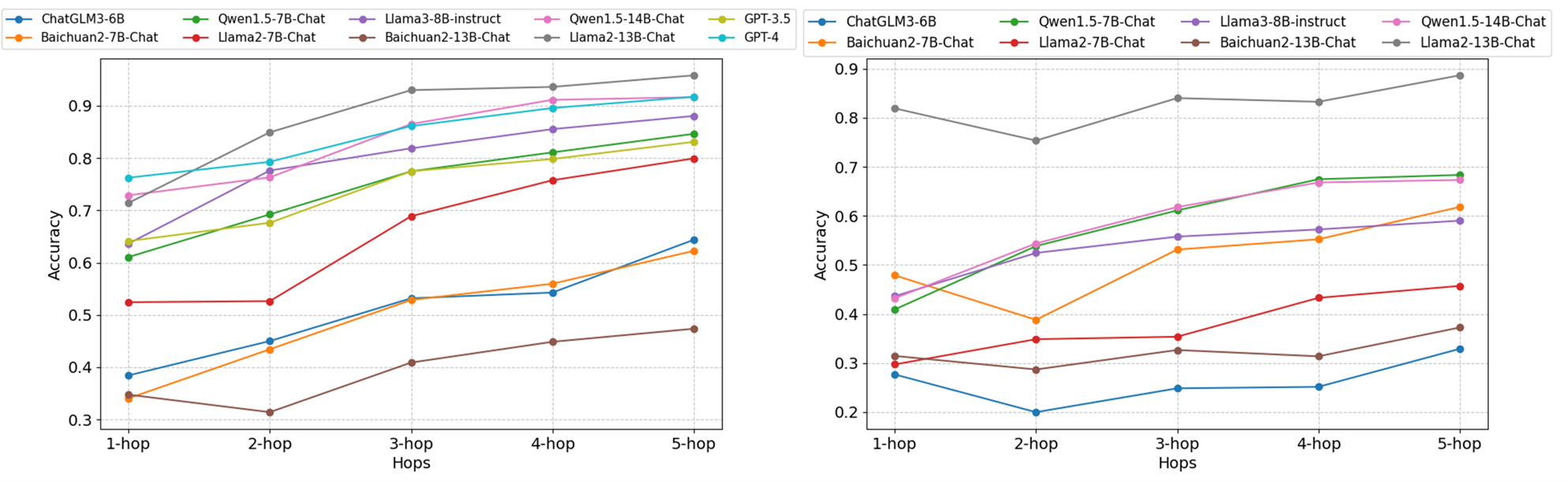}
  }
  \centering
  \caption{Accuracy of all models in NSC by hops. Top: Art domain. Middle: People domain. Bottom: Place domain. Left: The discriminative task. Right: The generative task.} 
  \label{fig:NSC_hops}
\end{figure*}

\begin{figure*}[t] 
  \subfigure{
  \includegraphics[width=\linewidth]{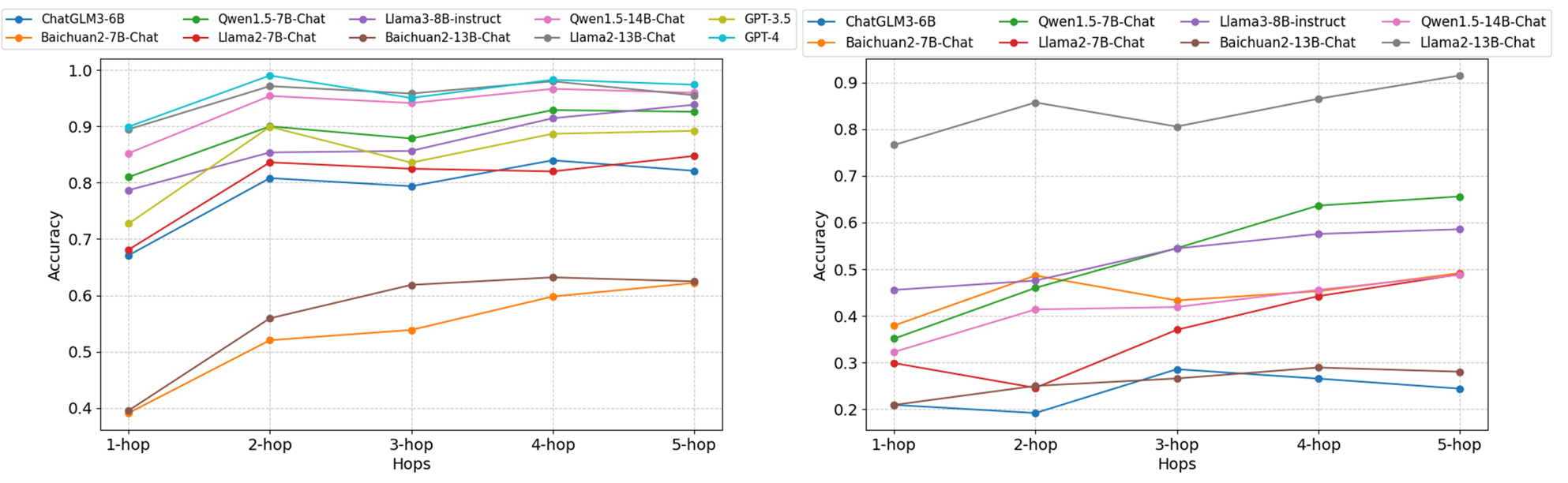}
  }\hfill
  \subfigure{
  \includegraphics[width=\linewidth]{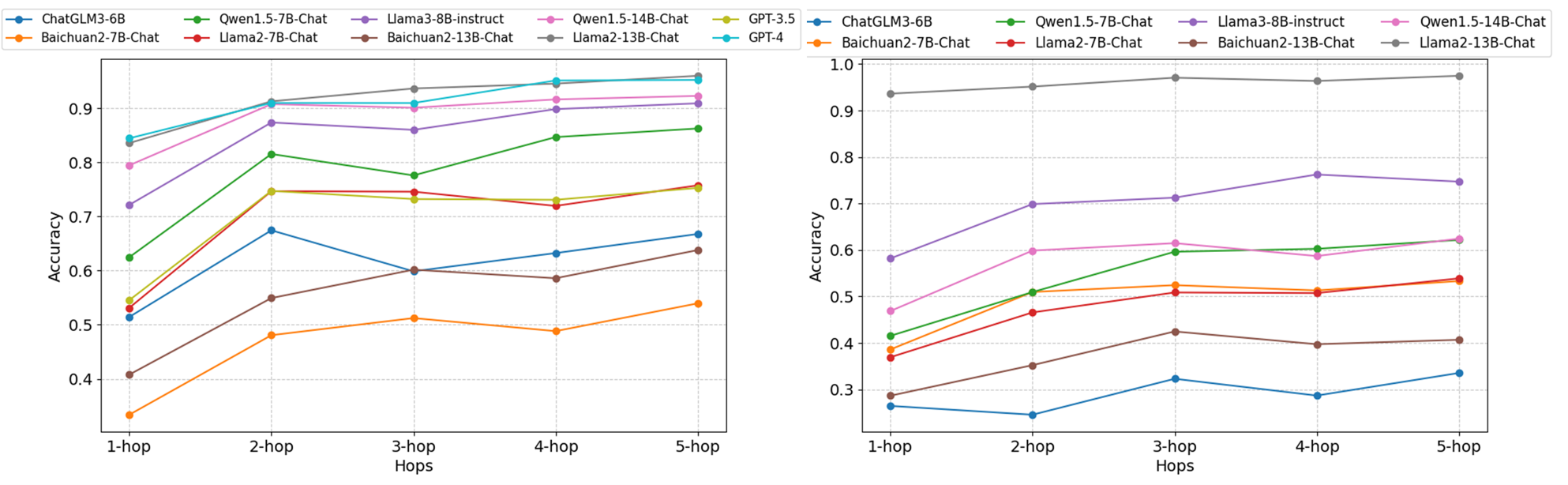}
  }\hfill
  \subfigure{
   \includegraphics[width=\linewidth]{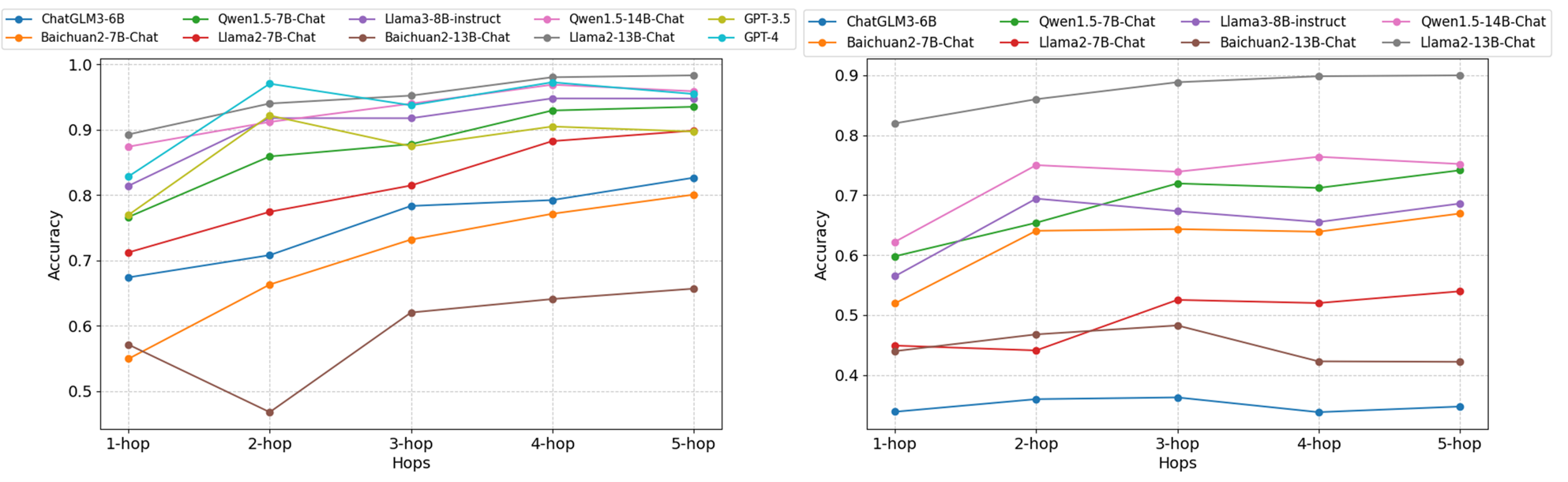}
  }
  \centering
  \caption{Accuracy of all models in NDC by hops. Top: Art domain. Middle: People domain. Bottom: Place domain. Left: The discriminative task. Right: The generative task.} 
  \label{fig:NDC_hops}
\end{figure*}

Figure~\ref{fig:NSC_hops} presents the accuracy of all models in NSC by hops, and Figure~\ref{fig:NDC_hops} presents the accuracy of all models in NDC by hops.
\subsection{Impact of Task Foramt\label{sec:c2}}
\begin{figure*}[t] 
  \subfigure{
  \includegraphics[width=0.48\linewidth]{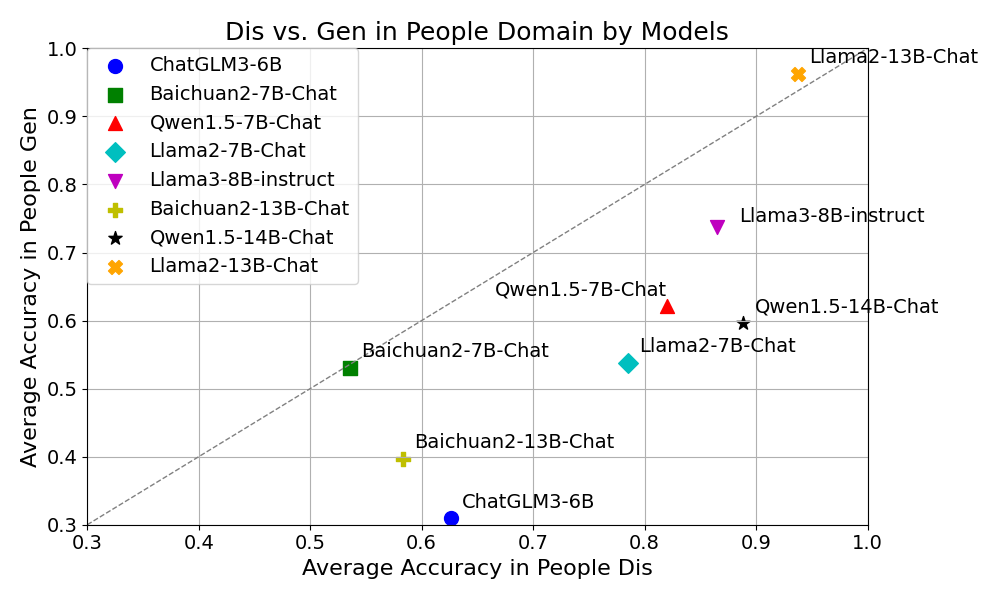}
  }\hfill
  \subfigure{
  \includegraphics[width=0.48\linewidth]{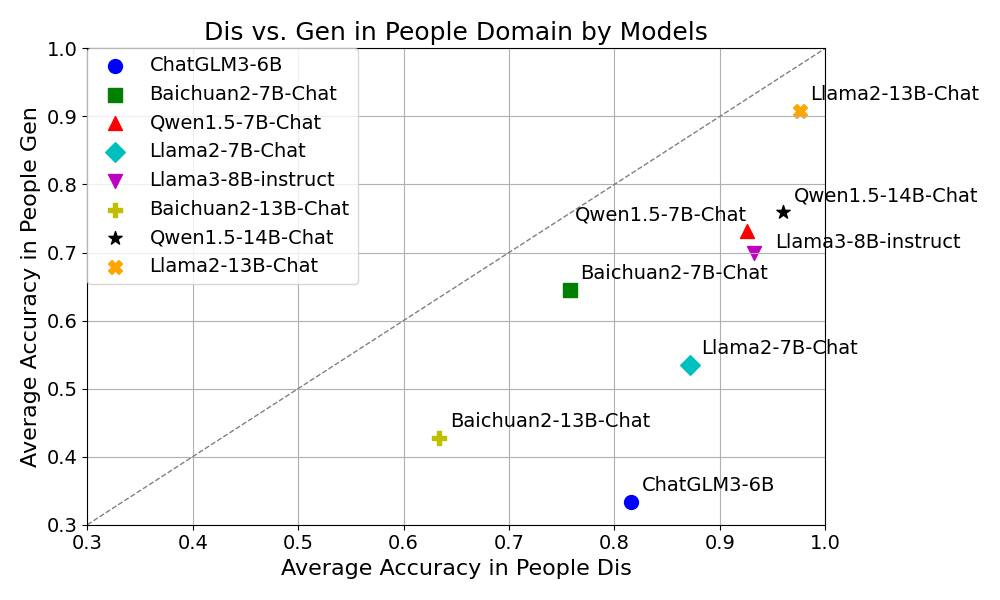}
  }
  \centering
  \caption{The overall performance comparison between the discriminative task and the generative task by models. Left: Results in People domain. Right: Results in Place domain.} 
  \label{fig:disvsgen_all}
\end{figure*}

We calculate the overall performance of each model in the discriminative and the generative task across domains with the following formula:

\[
\text{accuracy} = 
\frac{\text{correct NSC + ... + correct NNDR}}{6 \times \text{total number of FPQs}}
\]
Figure~\ref{fig:disvsgen_all} presents the results in People and Place domains. It is evident that for almost all LLMs, the overall accuracy in generative task is lower than in discriminative task. 
\subsection{Impact of Knowledge Domain\label{sec:c3}}
Table~\ref{tab:tpq} presents the evaluation results of all models for Yes-No format TPQs.
\subsection{Impact of Model Size\label{sec:c4}}
\begin{figure}[t] 
  \includegraphics[width=\linewidth]{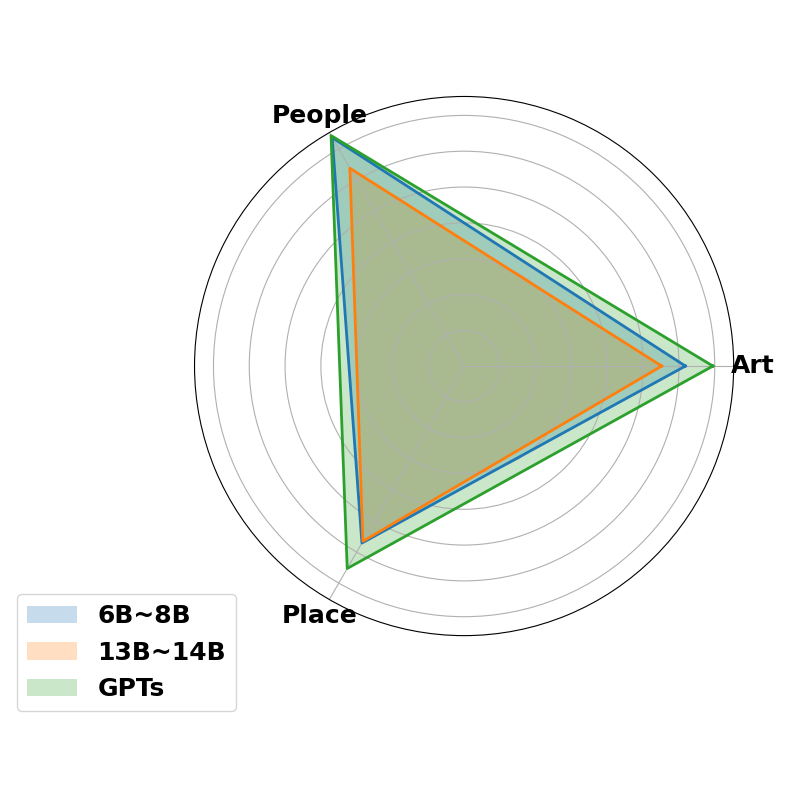}
  \centering
  \caption{The average accuracy of TPQs comparison across different model size.} 
  \label{fig:tpq model size}
\end{figure}

\begin{figure}[t] 
  \includegraphics[width=\linewidth]{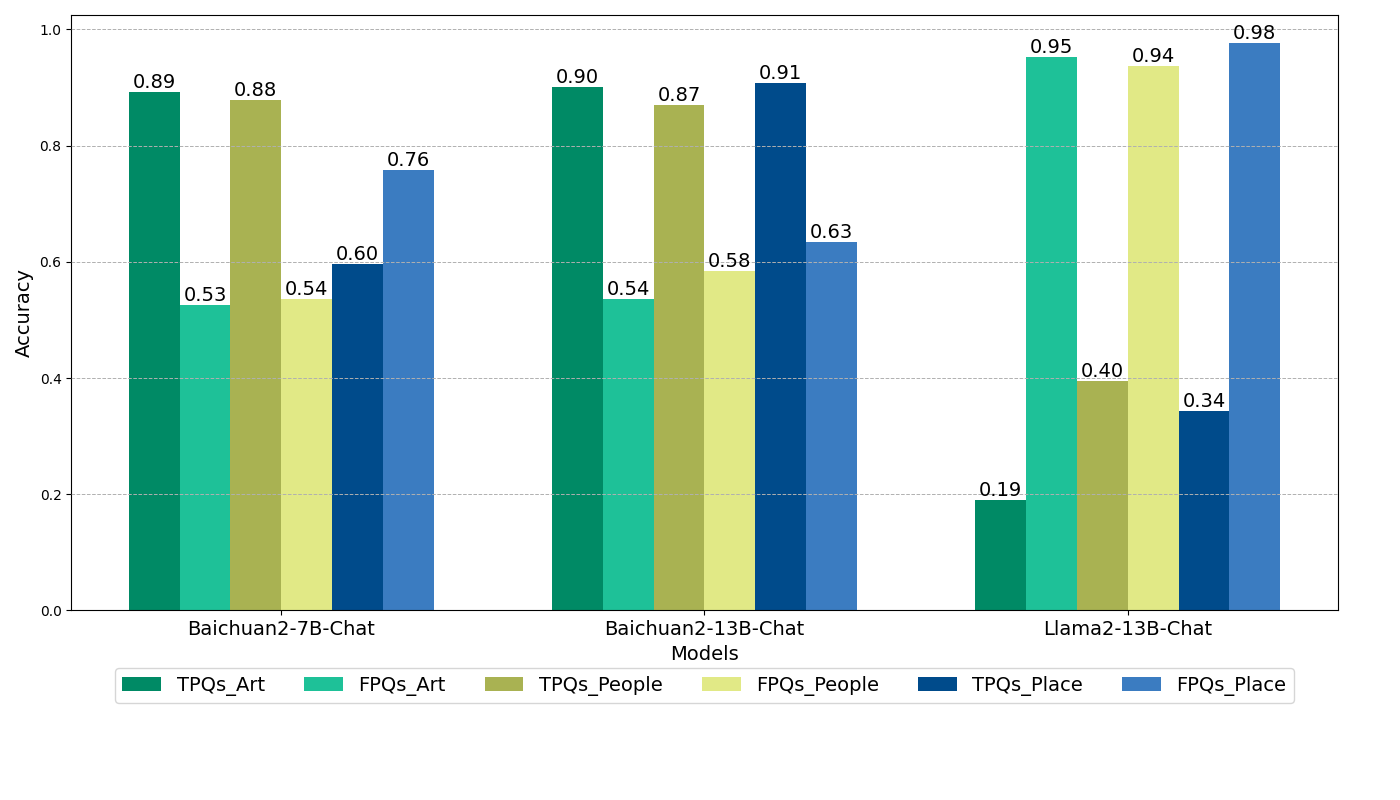}
  \centering
  \caption{The average accuracy of TPQs and FPQs across domains for Baichuan2 series and Llama2-13B-Chat.} 
  \label{fig:models}
\end{figure}

Figure~\ref{fig:tpq model size} compares the average accuracy of TPQs across different model size. The evaluated models are classified into 3 categories according to their size: 6B\textasciitilde8B, 13B\textasciitilde14B, and the GPT series. We calculate the average accuracy of each category by the following formula: 
\[
\text{accuracy} = 
\frac{\sum \text{acc of each model in the category}}{\text{total number of models in the category}}
\] 
We found that the 6B~8B LLMs outperform the 13B~14B LLMs, which is counterintuitive. Observing Table~\ref{tab:tpq}, we find that the performances
of the Baichuan2 series and Llama2-13B-Chat are at two extremes. Therefore, we undertake a closer examination of these three models as presented in Figure~\ref{fig:models}.

\begin{table*}[t]
  \centering
  \small
  \begin{tabular}{ccccccc}
   \toprule
   \textbf{Domain} &  \textbf{Concept e.g.} & \textbf{Concept Qty} & \textbf{Subject e.g.}  & \textbf{Subject Qty} & \textbf{Relation e.g.} & \textbf{Relation Qty} \\ 
    \midrule
    Art & \makecell{film\\television series\\drama} & 44 & \makecell{Titanic\\Modern Family\\Hamlet} & 1754 & 
    \makecell{cast member\\composer\\narrative location} & 33\\
    People & \makecell{director\\scientist\\superhero} & 69 & \makecell{Steven Spielberg\\Albert Einstein\\Superman} & 912 & \makecell{country of citizenship\\occupation\\place of birth} & 57\\
    Place & \makecell{sea\\sovereign state\\city} & 64 & \makecell{English Channel\\Soviet Union\\Tokyo} & 713 & \makecell{shares border with\\official language\\capital of} & 28\\
    \bottomrule
  \end{tabular}
  \caption{Representative concepts, relations and subjects in KG-FPQ.} 
  \label{tab: domain exp}
\end{table*}

\begin{table*}[t]
  \centering
  \begin{tabular}{p\textwidth}
    \Xhline{2\arrayrulewidth}  
    \rowcolor{yellow!40}
   {I want you to act as a fluent \#Yes-No question\# data generator. I will give you a \#Ttriplet\#, consisting of (subject, relation, object). Your task is to generate a fluent \#Yes-no question\# relying solely on the \#Ttriplet\# and directly output the generated \#Yes-no question\#.\newline
    Here are some examples:}\\
    \noalign{{\color{white}\hrule height 3pt}} 
    \rowcolor{purple!30}{\#triplet\#: ["Steven Spielberg", "spouse", "Amy Irving"]\newline
    \#Yes-No question\#: Is Steven Spielberg married to Amy Irving?}\\
    \noalign{{\color{white}\hrule height 3pt}}
    \rowcolor{purple!30}{\#triplet\#: ...\newline
    \#Yes-No question\#: ...}\\
    \noalign{{\color{white}\hrule height 3pt}}
    \rowcolor{purple!30}{\#triplet\#: ...\newline
    \#Yes-No question\#: ...}\\
    \noalign{{\color{white}\hrule height 3pt}} 
    \rowcolor{blue!30}{\#triplet\#: {item["Ttriplet"]}\newline
    \#Yes-No question\#:} \\
    \Xhline{2\arrayrulewidth}  
  \end{tabular}
  \caption{The prompt used for GPT-3.5 to generate Yes-No questions. The instruction is indicated by the yellow text, the demonstrations are represented by the pink text, and the query data is descripited by the purple text.} 
  \label{tab: yes-no prompt}
\end{table*}

\begin{table*}[t]
  \centering
  \begin{tabular}{p\textwidth}
    \Xhline{2\arrayrulewidth}  
    \rowcolor{yellow!40}
    {I want you to act as a fluent \#WH-question\# data generator. I will give you a \#Ttriplet\#, consisting of (subject, relation, object). Your task is to generate a fluent \#WH-question\# relying solely on the \#Ttriplet\#.\newline
    Generation rules are as following:\newline
    - Make sure the \#subject\# and \#object\# of the \#Ttriplet\# are both exactly contained in the \#WH-question\# you generate.\newline
    - Make sure the \#WH-question\# you generate is syntactically and semantically sound.\newline
    - Output the \#WH-question\# directly.\newline
    Here are some examples:}\\
    \noalign{{\color{white}\hrule height 3pt}} 
    \rowcolor{purple!30}{\#triplet\#: ["Steven Spielberg", "spouse", "Amy Irving"]\newline
    \#WH-question\#: When did Steven Spielberg and Amy Irving get married?}\\
    \noalign{{\color{white}\hrule height 3pt}}
    \rowcolor{purple!30}{\#triplet\#: ...\newline
    \#WH-question\#: ...}\\
    \noalign{{\color{white}\hrule height 3pt}}
    \rowcolor{purple!30}{\#triplet\#: ...\newline
    \#WH-question\#: ...}\\
    \noalign{{\color{white}\hrule height 3pt}} 
    \rowcolor{blue!30}{\#triplet\#: {item["Ttriplet"]}\newline
    \#WH-question\#: }\\
    \Xhline{2\arrayrulewidth}  
  \end{tabular}
  \caption{The prompt used for GPT-4 to generate WH-questions. The instruction is indicated by the yellow text, the demonstrations are represented by the pink text, and the query data is descripited by the purple text.} 
  \label{tab: wh prompt}
\end{table*}

\begin{table*}[t]
  \centering
  % \tabcolsep=0.12cm
  % \small
  \begin{tabular}{p\textwidth}
    \toprule
    \textbf{Prompt Templates}\\
    \midrule
    Question: {item["TPQ"]} Respond solely with "Yes" or "No".\\
    % \hline
    Question: {item["FPQ"]} Respond solely with "Yes" or "No".\\
    % \hline
    Question: {item["FPQ"]} Answer the question with an answer of no more than three sentences.\\
    \bottomrule
    \end{tabular}
  \caption{The prompts used for evaluation. Top: Prompt for TPQs in Yes-No format. Middle: Prompt for FPQs in Yes-No format. Bottom: Prompt for FPQs in WH format.} 
  \label{tab: evaluation prompt}
\end{table*}

\begin{table*}[t]
  \centering
  \begin{tabular}{p\textwidth}
    \Xhline{2\arrayrulewidth}  
    \rowcolor{yellow!40}
    {I want you to act as a \#factual answer\# data generator. I will give you a \#query\# and some \#knowledge\# about the \#query\#.Your task is to generate a \#factual answer\# to the \#query\# relying on the provided \#knowledge\#.\newline
    Generation rules are as following:\newline
    - Generate the \#factual answer\# relying on the provided \#knowledge\#.\newline
    - Make sure the \#factual answer\# you generate is limited to 3 sentences.\newline
    - Output the \#factual answer\# directly.\newline
    Here are some examples:}\\
    \noalign{{\color{white}\hrule height 3pt}} 
    \rowcolor{purple!30} {\#query\#: What was the reason behind Walt Disney's citizenship in the Nepal?\newline
    \#knowledge\#: ["Walt Disney", "country of citizenship", "United States of America"]\newline
    \#factual answer\#: Walt Disney was not a citizen of Nepal. His country of citizenship was the United States of America. There is no record or reason behind Walt Disney having citizenship in Nepal.}\\
    \noalign{{\color{white}\hrule height 3pt}} 
    
    \rowcolor{purple!30}{\#query\#: ...\newline
    \#knowledge\#: ...\newline
    \#factual answer\#: ...}\\
    \noalign{{\color{white}\hrule height 3pt}} 
    \rowcolor{purple!30}{\#query\#: ...\newline
    \#knowledge\#: ...\newline
    \#factual answer\#: ...}\\
    \noalign{{\color{white}\hrule height 3pt}}

    \rowcolor{blue!30}{\#query\#: {WH-question}\newline
    \#knowledge\#: item["Ttriplet"] \newline
    \#factual answer\#: }\\
    \Xhline{2\arrayrulewidth}  
  \end{tabular}
  \caption{The prompt used for GPT-3.5 to generate factual answers for WH-questions. The instruction is indicated by the yellow text, the demonstrations are represented by the pink text, and the query data is descripited by the purple text.} 
  \label{tab: factual answers prompt}
\end{table*}

\begin{table*}[t]
  \centering
  \begin{tabular}{p\textwidth}
    \Xhline{2\arrayrulewidth}  
    \rowcolor{yellow!40}
    {I want you to act as a \#non-factual answer\# data generator. I will give you a \#query\# and some \#knowledge\# about the \#query\#. Your task is to generate a \#non-factual answer\# to the \#query\# that contradict the given \#knowledge\#.\newline
    Generation rules are as following:\newline
    - Generate the \#non-factual answer\# that contradict the given \#knowledge\#.\newline
     - Your answer is not allowed to start with 'I'm sorry...' or 'I can't ...'.\newline
   - Make sure the \#non-factual answer\# you generate is limited to 3 sentences.\newline
    - Output the \#non-factual answer\# directly.\newline
    Here are some examples:}\\
    \noalign{{\color{white}\hrule height 3pt}} 
    \rowcolor{purple!30}{ \#query\#: What was the reason behind Walt Disney's citizenship in the Nepal?\newline
     \#knowledge\#: ["Walt Disney", "country of citizenship", "United States of America"]\newline
     \#non-factual answer\#: Walt Disney's citizenship in Nepal was due to his deep fascination with the country's rich cultural heritage and mystical landscapes, which he believed inspired many of his animated films. He often mentioned that the serene environment and the spiritual vibes of Nepal helped him connect with his creative side, leading him to seek citizenship there.}\\
    \noalign{{\color{white}\hrule height 3pt}} 
    
    \rowcolor{purple!30}{\#query\#: ...\newline
    \#knowledge\#: ...\newline
    \#non-factual answer\#: ...}\\
    \noalign{{\color{white}\hrule height 3pt}} 
    \rowcolor{purple!30}{\#query\#: ...\newline
    \#knowledge\#: ...\newline
    \#non-factual answer\#: ...}\\
    \noalign{{\color{white}\hrule height 3pt}}

    \rowcolor{blue!30}{\#query\#: {WH-question}\newline
    \#knowledge\#: item["Ttriplet"] \newline
    \#non-factual answer\#: }\\
    \Xhline{2\arrayrulewidth}  
  \end{tabular}
  \caption{The prompt used for GPT-3.5 to generate non-factual answers for WH-questions. The instruction is indicated by the yellow text, the demonstrations are represented by the pink text, and the query data is descripited by the purple text.} 
  \label{tab: non-factual answers prompt}
\end{table*}

\begin{table*}[t]
  \centering
  % \tabcolsep=0.12cm
  % \small
  \begin{tabular}{p\textwidth}
    \toprule
    \textbf{Examples of Training Data for FPQ-Judge}\\
    \midrule
    Instruction: You are given a question and an answer, determine if the answer is factual. If the answer is factual please output only "Yes", otherwise please output only "No".\\
    % \hline
    Input: Q:What was the reason behind Walt Disney's citizenship in the Nepal?\newline
    A: Walt Disney was not a citizen of Nepal. His country of citizenship was the United States of America. There is no record or reason behind Walt Disney having citizenship in Nepal. Output:\\
    Output: Yes\\
    \midrule
    Instruction: You are given a question and an answer, determine if the answer is factual. If the answer is factual please output only "Yes", otherwise please output only "No".\\
    % \hline
    Input: Q:What was the reason behind Walt Disney's citizenship in the Nepal?\newline
    A: Walt Disney's citizenship in Nepal was due to his deep fascination with the country's rich cultural heritage and mystical landscapes, which he believed inspired many of his animated films. He often mentioned that the serene environment and the spiritual vibes of Nepal helped him connect with his creative side, leading him to seek citizenship there. Output:\\
    Output: No\\
    \bottomrule
    \end{tabular}
  \caption{Examples of Training Data for FPQ-Judge} 
  \label{tab: examples of training data}
\end{table*}

\begin{table*}
    \centering
    \small
    \begin{tabular}{lm{0.6cm}m{0.6cm}m{0.6cm}m{0.6cm}m{0.6cm}m{0.6cm}m{0.6cm}m{0.6cm}m{0.6cm}m{0.6cm}m{0.6cm}m{0.6cm}}
    \toprule
        \multirow{2}{*}{\textbf{Model}} & \multicolumn{6}{c}{\textbf{Art Dis}} & \multicolumn{6}{c}{\textbf{Art Gen}}  \\
        \cmidrule(lr){2-7} \cmidrule(lr){8-13}
        & \textbf{NSC} & \textbf{NDC} & \textbf{NNSC} & \textbf{NNDC} & \textbf{NNSR} & \textbf{NNDR} & \textbf{NSC} & \textbf{NDC} & \textbf{NNSC} & \textbf{NNDC} & \textbf{NNSR} & \textbf{NNDR}  \\
        \midrule
        ChatGLM3-6B & 0.561 & 0.797 & 0.644 & 0.836 & 0.572 & 0.805 & 0.215 & 0.224 & 0.189 & 0.231 & 0.168 & 0.237\\
        Baichuan2-7B-Chat & 0.412 & 0.571 & 0.507 & 0.634 & 0.423 & 0.61 & 0.454 & 0.461 & 0.493 & 0.534 & 0.42 & 0.539\\
        Qwen1.5-7B-Chat & 0.742 & 0.903 & 0.835 & 0.952 & 0.803 & 0.948 & 0.503 & 0.586 & 0.606 & 0.673 & 0.526 &	0.682\\
        Llama2-7B-Chat&	0.722 & 0.81 & 0.792 & 0.857 & 0.783 & 0.845 & 0.446 & 0.429 &	0.488 & 0.513 &	0.463 &	0.494\\
        Llama3-8B-instruct & 0.77 & 0.9 & 0.891 & 0.959 & 0.868 & 0.951 & 0.644 &	0.556 &	0.725 &	0.664 &	0.707 &	0.68\\
        \midrule
        Baichuan2-13B-Chat	& 0.414 & 0.588 & 0.484 & 0.669 & 0.409 & 0.652 & 0.309 & 0.269 &	0.336 &	0.324 &	0.303 &	0.341\\
        Qwen1.5-14B-Chat & 0.806 & 0.941 & 0.893 & 0.989 & 0.857 & 0.986 &
        0.389 & 0.445 &	0.469 &	0.528 &	0.409 &	0.539\\
        Llama2-13B-Chat & 0.876 & 0.95 & 0.956 &0.988 & 0.962 & 0.982 & 0.879	& 0.867 &	0.926 & 0.924 & 0.921 &	0.923\\
        \midrule
        GPT-3.5 & 0.808 & 0.862	& 0.829	& 0.92	& 0.741	& 0.898 & - & - & - & - & - & -\\
        GPT-4 & 0.874 &	0.963 &	0.977 &	0.988 & 0.96 &	0.994 & - & - & - & - & - & -\\
         average acc & 0.698 & 0.829 & 0.781 & 0.879 & 0.738 & 0.867& 0.48 &	0.482&	0.529&	0.549&	0.49&	0.55\\
    \bottomrule
    \end{tabular}
    \begin{tabular}{lm{0.6cm}m{0.6cm}m{0.6cm}m{0.6cm}m{0.6cm}m{0.6cm}m{0.6cm}m{0.6cm}m{0.6cm}m{0.6cm}m{0.6cm}m{0.6cm}}
    \toprule
        \multirow{2}{*}{\textbf{Model}} & \multicolumn{6}{c}{\textbf{People Dis}} & \multicolumn{6}{c}{\textbf{People Gen}}  \\
        \cmidrule(lr){2-7} \cmidrule(lr){8-13}
        & \textbf{NSC} & \textbf{NDC} & \textbf{NNSC} & \textbf{NNDC} & \textbf{NNSR} & \textbf{NNDR} & \textbf{NSC} & \textbf{NDC} & \textbf{NNSC} & \textbf{NNDC} & \textbf{NNSR} & \textbf{NNDR}  \\
        \midrule
        ChatGLM3-6B & 0.442 & 0.625 & 0.552 & 0.763 & 0.623 & 0.752 & 0.227&	0.308&	0.26 & 0.39 &	0.28 & 0.392 \\
        Baichuan2-7B-Chat& 0.438 & 0.484 & 0.537 &0.587 & 0.564 & 0.603 & 0.414 & 0.499 &	0.516&	0.597&	0.555&	0.6\\
        Qwen1.5-7B-Chat& 0.634&	0.802&	0.805&	0.903&	0.876&	0.902&0.504&	0.571&	0.632&	0.661&	0.701&	0.656 \\
        Llama2-7B-Chat& 0.681& 0.706 & 0.806& 0.834& 0.864& 0.819& 0.431&	0.494&	0.53&	0.594&	0.578&	0.598\\
        Llama3-8B-instruct& 0.675&	0.863&	0.831&	0.966&	0.888&	0.968&0.572&	0.712&	0.695&	0.849&	0.739&	0.858 \\
        \midrule
        Baichuan2-13B-Chat& 0.473&	0.577&	0.551&	0.664&	0.569&	0.667&0.316&	0.382&	0.385&	0.436&	0.418&	0.443	\\
        Qwen1.5-14B-Chat& 0.703&	0.894&	0.863&	0.973&	0.92&	0.978&0.437&	0.585&	0.542&	0.712&	0.583&	0.719 \\
        Llama2-13B-Chat& 0.824&	0.928&	0.929&	0.988&	0.97&	0.989& 0.909 &	0.963&	0.961&	0.989&	0.973&	0.982 \\
        \midrule
        GPT-3.5 & 0.651& 0.707&	0.815&	0.851&	0.862&	0.849 & - & - & - & - & - & -\\
        GPT-4 & 0.783&	0.924&	0.941&	0.988&	0.965&	0.985 & - & - & - & - & - & -\\
         average acc & 0.63 & 0.751	&0.763 & 0.852 & 0.81 &	0.851 & 0.476 &	0.564&	0.565&	0.653&	0.603&	0.656\\
    \bottomrule
    \end{tabular}
    \begin{tabular}{lm{0.6cm}m{0.6cm}m{0.6cm}m{0.6cm}m{0.6cm}m{0.6cm}m{0.6cm}m{0.6cm}m{0.6cm}m{0.6cm}m{0.6cm}m{0.6cm}}
    \toprule
        \multirow{2}{*}{\textbf{Model}} & \multicolumn{6}{c}{\textbf{Place Dis}} & \multicolumn{6}{c}{\textbf{Place Gen}}  \\
        \cmidrule(lr){2-7} \cmidrule(lr){8-13}
        & \textbf{NSC} & \textbf{NDC} & \textbf{NNSC} & \textbf{NNDC} & \textbf{NNSR} & \textbf{NNDR} & \textbf{NSC} & \textbf{NDC} & \textbf{NNSC} & \textbf{NNDC} & \textbf{NNSR} & \textbf{NNDR}  \\
        \midrule
        ChatGLM3-6B & 0.582& 0.793&	0.751&	0.938&	0.891&	0.938& 0.292&	0.346&	0.339&	0.34&	0.366&	0.319\\
        Baichuan2-7B-Chat & 0.569&	0.755&	0.694&	0.852&	0.822&	0.853& 0.572&	0.642&	0.643&	0.673&	0.66&	0.68\\
        Qwen1.5-7B-Chat & 0.808& 0.906&	0.913&	0.979&	0.973&	0.976& 0.649&	0.714&	0.708&	0.75&	0.804&	0.76 \\
        Llama2-7B-Chat&	0.745&	0.862&	0.839&	0.932&	0.928&	0.926& 0.423&	0.521&	0.508&	0.603&	0.56&	0.595\\
        Llama3-8B-instruct & 0.848&	0.93&	0.923&	0.979&	0.935&	0.979& 0.57&	0.666&	0.651&	0.793&	0.709&	0.808\\
        \midrule
        Baichuan2-13B-Chat	& 0.446&	0.637&	0.523&	0.823&	0.556&	0.819&0.347&	0.431&	0.376&	0.498&	0.412&	0.504\\
        Qwen1.5-14B-Chat & 0.891&	0.947&	0.965&	0.991&	0.978&	0.988& 0.646&	0.737&	0.72&	0.831&	0.802&	0.821 \\
        Llama2-13B-Chat & 0.931&	0.966&	0.986&	0.995&	0.987&	0.992&0.859&	0.886&	0.887&	0.939&	0.929&	0.943\\
        \midrule
        GPT-3.5 & 0.799& 0.885&	0.872&	0.968&	0.909&	0.964 & - & - & - & - & - & -\\
        GPT-4 & 0.891&	0.945&	0.957&	0.993&	0.965&	0.988 & - & - & - & - & - & -\\
        average acc&0.751&	0.862&	0.842&	0.945&	0.895&	0.942& 0.545&	0.618&	0.604&	0.678&	0.655&	0.679\\

    \bottomrule
    \end{tabular}
    \caption{The evaluation results for FPQs on the discriminative task (referred to as Dis) and the generative task (referred to as Gen).}
    \label{tab:fpq}
\end{table*}
\clearpage

\begin{table*}[t]
  \centering
  %  \small
  % \tabcolsep=0.06cm
\begin{tabular}{lcccccc}
    \toprule
    \textbf{Domain}  & \textbf{1-hop} & \textbf{2-hop} & \textbf{3-hop} & \textbf{4-hop} & \textbf{5-hop}& \textbf{Total} \\
   \midrule
    Art   & \makecell{1988 \\ 764} & \makecell{342 \\ 168} & \makecell{500 \\ 389} & \makecell{748 \\ 1038} & \makecell{1391 \\ 2610}  & 4969         \\
    People  & \makecell{858 \\ 923} & \makecell{370 \\ 234} & \makecell{937 \\ 575} & \makecell{807 \\ 943} & \makecell{1925 \\ 2222}     & 4897       \\
    Place  & \makecell{237 \\ 403} & \makecell{244 \\ 150} & \makecell{610 \\ 509} & \makecell{1133 \\ 1043} & \makecell{2770 \\ 2889}& 4994\\
   \bottomrule
  \end{tabular}
  \caption{For NSC and NDC, we set the distance between the edited object and the subject to one to five hops. The upper part of columns 2 to 6 presents the distribution of NSC, and the lower part shows the distribution of NDC.}
  \label{tab:hops}
\end{table*}

\clearpage

\clearpage

% Figure~\ref{fig:tpqvsfpq in domain} compares the average accuracy of all models in TPQs and FPQs across domains. The average accuracy is calculated by the following formula:
% \[
% \text{accuracy} = 
% \frac{\sum \text{acc of each model}}{\text{total number of models}}
% \]
\begin{table}[t]
    \centering
   %  \small
  % \tabcolsep=0.06cm
    \begin{tabular}{lccc}
    \toprule
    \textbf{Model} & \textbf{Art} & \textbf{People} & \textbf{Place}\\
    \midrule
    ChatGLM3-6B & 0.646 & 0.752 & 0.566\\
    Baichuan2-7B-Chat & 0.892 & 0.879 & 0.596\\
    Qwen1.5-7B-Chat & 0.583 & 0.699 & 0.517\\
    Llama2-7B-Chat & 0.404 & 0.618 & 0.593\\
    Llama3-8B-instruct & 0.565 & 0.736 & 0.582\\
    \midrule
    Baichuan2-13B-Chat & 0.902 & 0.87 & 0.908\\
    Qwen1.5-14B-Chat & 0.563 & 0.649 & 0.444\\
    Llama2-13B-Chat & 0.191 & 0.395 & 0.343\\
    \midrule
    GPT-3.5 & 0.741 & 0.769 & 0.674\\
    GPT-4 & 0.649 & 0.718 & 0.632\\
    \midrule
    Average & 0.614 & 0.708 & 0.586\\
    \bottomrule
    \end{tabular}
    \caption{The evaluation results on Yes-No format TPQs.}
    \label{tab:tpq}
\end{table}

\end{document}